\documentclass[11pt,a4paper]{article}
\usepackage[hyperref]{emnlp2020}
\usepackage{times}
\usepackage{latexsym}
\usepackage{graphicx}
\usepackage{url}
\usepackage{booktabs, multirow}
\usepackage{xspace}
\usepackage{subcaption}
\usepackage{microtype}
\usepackage{adjustbox}
\usepackage{color, colortbl}
\usepackage{amsmath}
\usepackage{tabularx}
\usepackage[utf8]{inputenc}
\usepackage{arydshln}
\usepackage{todonotes}
\usepackage{multirow}
\usepackage{mdwlist}

\graphicspath{{./images/}}

\newcommand{\en}{{\textsc{en}}\xspace}
\newcommand{\es}{{\textsc{es}}\xspace}
\newcommand{\qu}{{\textsc{qu}}\xspace}
\newcommand{\eu}{{\textsc{eu}}\xspace}
\newcommand{\gl}{{\textsc{gl}}\xspace}
\newcommand{\ta}{{\textsc{ta}}\xspace}
\newcommand{\ur}{{\textsc{ur}}\xspace}
\newcommand{\bn}{{\textsc{bn}}\xspace}
\newcommand{\hf}{{\textsc{HFREQ}}\xspace}
\newcommand{\mf}{{\textsc{MFREQ}}\xspace}
\newcommand{\lf}{{\textsc{LFREQ}}\xspace}
\newcommand{\af}{{\textsc{all-FREQ}}\xspace}
\newcommand{\elltwo}{{\textsc{L2}}\xspace}
\newcommand{\mc}{{\textsc{mc}}\xspace}

\definecolor{Gray}{gray}{0.92}

\newcolumntype{Y}{>{\centering\arraybackslash}X}

\aclfinalcopy 

\title{Are All Good Word Vector Spaces Isomorphic?}

\author{Ivan Vuli\'c$^{\mathbf{1}}$\thanks{{ } All authors contributed equally to this work.} ~~ Sebastian Ruder$^{\mathbf{2}}$\footnotemark[1] ~~ Anders S{\o}gaard$^{\mathbf{3,4}}$\footnotemark[1] \\
$^{\mathbf{1}}$ Language Technology Lab, University of Cambridge \\
$^{\mathbf{2}}$ DeepMind \\
$^{\mathbf{3}}$ Department of Computer Science, University of Copenhagen \\
$^{\mathbf{4}}$ Google Research, Berlin \\
\texttt{iv250@cam.ac.uk} \hspace{0.8em} \texttt{ruder@google.com} \hspace{0.8em} \texttt{soegaard@di.ku.dk} }

\date{}

\begin{document}
\maketitle
\begin{abstract}

Existing algorithms for aligning cross-lingual word vector spaces assume that vector spaces are approximately isomorphic. As a result, they perform poorly or fail completely on non-isomorphic spaces. Such non-isomorphism has been hypothesised to result from typological differences between languages. In this work, we ask whether non-isomorphism is also crucially {\em a sign of degenerate word vector spaces}. We present a series of experiments across diverse languages which show that variance in performance across language pairs is not only due to typological differences, but can mostly be attributed to the size of the monolingual resources available, and to the properties and duration of monolingual training (e.g. ``under-training'').


\end{abstract}



\section{Introduction}
\label{s:introduction}
Word embeddings have been argued to reflect how language users organise concepts \cite{Mandera:2017,Asr:2018naacl}. The extent to which they really do so has been evaluated, e.g., using semantic word similarity and association norms \cite{hill2015simlex,Gerz:2016emnlp}, and word analogy benchmarks \cite{Mikolov2013naacl}. If word embeddings reflect more or less language-independent conceptual organisations, word embeddings in different languages can be expected to be near-isomorphic. Researchers have exploited this to learn linear transformations between such spaces \cite{Mikolov2013exploiting,Glavas2019}, which have been used to induce bilingual dictionaries, as well as to facilitate multilingual modeling and cross-lingual transfer \cite{Ruder2019survey}.


In this paper, we show that near-isomorphism arises only with sufficient amounts of training. This is of practical interest for applications of linear alignment methods for cross-lingual word embeddings. It furthermore provides us with an explanation for reported failures to align word vector spaces in different languages \cite{sogaard2018limitations,Artetxe2018}, which has so far been largely attributed \textit{only} to inherent typological differences.

\begin{figure}[!t]
    \centering
    \includegraphics[width=0.96\linewidth]{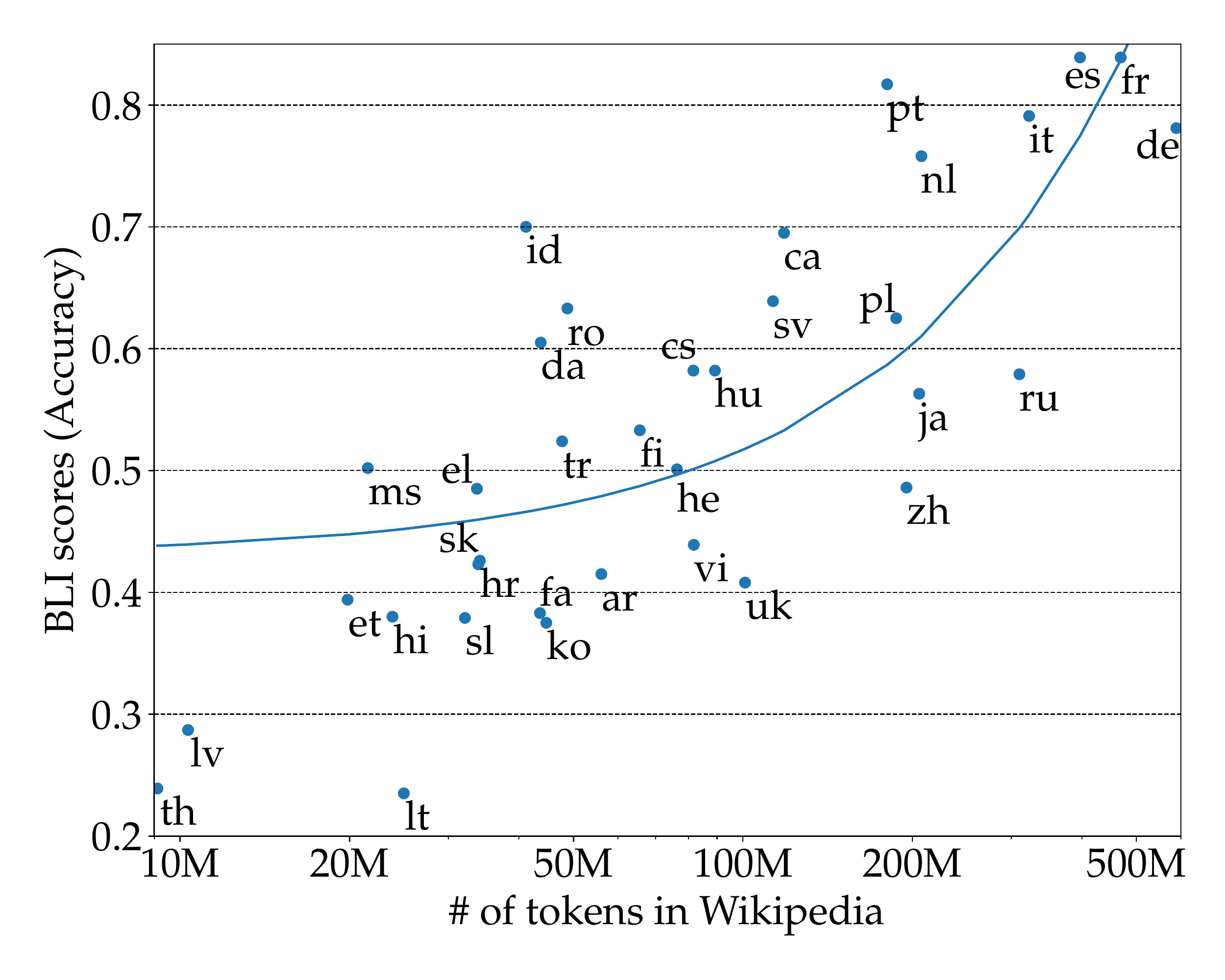}
    \vspace{-1.5mm}
    \caption{Performance of a state-of-the-art BLI model mapping from English to a target language and the size of the target language Wikipedia are correlated. Linear fit shown as a blue line (log scale).}
    \label{fig:token-counts-vs-scores}
    \vspace{-1.5mm}
\end{figure}

In fact, the amount of data used to induce the monolingual embeddings is predictive of the quality of the aligned cross-lingual word embeddings, as evaluated on bilingual lexicon induction (BLI). Consider, for motivation, Figure~\ref{fig:token-counts-vs-scores}; it shows the performance of a state-of-the-art alignment method---RCSLS with iterative normalisation \cite{Zhang2019}---on mapping English embeddings onto embeddings in other languages, and its correlation ($\rho = 0.72$) with the size of the tokenised target language Polyglot Wikipedia \cite{polyglot:2013:ACL-CoNLL}. 

We investigate to what extent the amount of data available for some languages and corresponding training conditions provide a {\em sufficient}~explanation for the variance in reported results; that is, whether it is the full story or not: The answer is 'almost', that is, its interplay with inherent typological differences does have a crucial impact on the `alignability' of monolingual vector spaces.

We first discuss current standard methods of quantifying the degree of near-isomorphism between word vector spaces (\S\ref{s:quantify-isomorphism}). We then outline training settings that may influence isomorphism (\S\ref{s:isomorphism-and-learning}) and present a novel experimental protocol for learning cross-lingual word embeddings that simulates a low-resource environment, and also controls for topical skew and differences in morphological complexity (\S\ref{s:simulated-low-resource}). We focus on two groups of languages: \textbf{1)} Spanish, Basque, Galician, and Quechua, and \textbf{2)} Bengali, Tamil, and Urdu, as these are arguably spoken in culturally related regions, but have very different morphology. Our experiments, among other findings, indicate that a low-resource version of Spanish is as difficult to align to English as Quechua, challenging the assumption from prior work that the primary issue to resolve in cross-lingual word embedding learning is language dissimilarity (instead of, e.g., procuring additional raw data for embedding training). We also show that by controlling for different factors, we reduce the gap between aligning Spanish and Basque to English from 0.291 to 0.129. Similarly, under these controlled circumstances, we do not observe any substantial performance difference between aligning Spanish and Galician to English, or between aligning  Bengali and Tamil to English.

We also investigate the learning dynamics of monolingual word embeddings and their impact on BLI performance and near-isomorphism of the resulting word vector spaces (\S\ref{s:experimental}), finding training duration, amount of monolingual resources, preprocessing, and self-learning all to have a large impact. The findings are verified across a set of typologically diverse languages, where we pair English with Spanish, Arabic, and Japanese. 

We will release our new evaluation dictionaries and subsampled Wikipedias controlling for topical skew and morphological differences to facilitate future research at: \url{https://github.com/cambridgeltl/iso-study}.






\section{Isomorphism of Vector Spaces}
\label{s:isomorphism}
Studies analyzing the qualities of monolingual word vector spaces have focused on intrinsic tasks \cite{Baroni2014}, correlations \cite{Tsvetkov2015}, and subspaces \cite{Yaghoobzadeh2016}. In the cross-lingual setting, the most important indicator for performance has been the \emph{degree of isomorphism}, that is, how (topologically) similar the structures of the two vector spaces are. 

\vspace{1.8mm}
\noindent \textbf{Mapping-based approaches} The prevalent way to learn a cross-lingual embedding space, especially in low-data regimes, is to learn a mapping between a source and a target embedding space \cite{Mikolov2013exploiting}. Such mapping-based approaches assume that the monolingual embedding spaces are isomorphic, i.e., that one can be transformed into the other via a linear transformation \cite{Xing2015orthogonal,Artetxe2018}. Recent unsupervised approaches rely even more strongly on this assumption: They assume that the structures of the embedding spaces are so similar that they can be aligned by minimising the distance between the transformed source language and the target language embedding space \cite{Zhang2017earthmover,Conneau2018,Xu2018,Alvarez-Melis2018gromov,Hartmann:2019neurips}.

\subsection{Quantifying Isomorphism} \label{s:quantify-isomorphism}

We employ measures that quantify isomorphism in three distinct ways---based on graphs, metric spaces, and vector similarity.

\vspace{1.8mm}
\noindent \textbf{Eigenvector similarity \cite{sogaard2018limitations}} Eigenvector similarity (EVS) estimates the degree of isomorphism based on properties of the nearest neighbour graphs of the two embedding spaces. We first length-normalise embeddings in both embedding spaces and compute the nearest neighbour graphs on a subset of the top most frequent $N$ words. We then calculate the Laplacian matrices $L_1$ and $L_2$ of each graph. For $L_1$, we find the smallest $k_1$ such that the sum of its $k_1$ largest eigenvalues $\sum^{k_1}_{i=1} {\lambda_1}_i$ is at least 90\% of the sum of all its eigenvalues. We proceed analogously for $k_2$ and set $k=\min (k_1, k_2)$. The eigenvector similarity metric $\Delta$ is now the sum of the squared differences of the $k$ largest Laplacian eigenvalues: $\Delta = \sum^k_{i=1}({\lambda_1}_i - {\lambda_2}_i)^2$. The lower $\Delta$, the \emph{more} similar are the graphs and the more isomorphic are the embedding spaces.

\vspace{1.8mm}
\noindent \textbf{Gromov-Hausdorff distance \cite{Patra2019}} The Hausdorff distance is a measure of the worst case distance between two metric spaces $\mathcal{X}$ and $\mathcal{Y}$ with a distance function $d$: 
%
{\normalsize
\begin{equation*}
\begin{split}
    \mathcal{H}(\mathcal{X}, \mathcal{Y}) = \max\{& \sup_{x \in \mathcal{X}} \inf_{y \in \mathcal{Y}} d(x, y), \\
    &\sup_{y \in \mathcal{Y}} \inf_{x \in \mathcal{X}} d(x, y)\}
\end{split}
\end{equation*}}%
Intuitively, it measures the distance between the nearest neighbours that are farthest apart. The Gromov-Hausdorff distance (GH) in turn minimises this distance over all isometric transforms (orthogonal transforms in our case as we apply mean centering) $\mathcal{X}$ and $\mathcal{Y}$ as follows:
{\normalsize
\begin{equation*}
    \mathcal{GH}(\mathcal{X}, \mathcal{Y}) = \inf_{f,g} \mathcal{H}(f(\mathcal{X}), g(\mathcal{Y}))
\end{equation*}}%
\noindent In practice, $\mathcal{GH}$ is calculated by computing the Bottleneck distance between the metric spaces \cite{chazal2009gromov,Patra2019}.

\vspace{1.8mm}
\noindent \textbf{Relational similarity} As an alternative, we consider a simpler measure inspired by \newcite{Zhang2019}. This measure, dubbed RSIM, is based on the intuition that the similarity distributions of translations within each language should be similar. We first take $M$ translation pairs $(m_s, m_t)$ from our bilingual dictionary. We then calculate cosine similarities for each pair of words $(m_s, n_s)$ on the source side where $m_s \neq n_s$ and do the same on the target side. Finally, we compute the Pearson correlation coefficient $\rho$ of the sorted lists of similarity scores. Fully isomorphic embeddings would have a correlation of $\rho=1.0$, and the correlation decreases with lower degrees of isomorphism.\footnote{There are other measures that quantify similarity between word vectors spaces based on network modularity \cite{Fujinuma2019}, singular values obtained via SVD \cite{Dubossarsky:2020emnlp}, and external resources such as sense-aligned corpora \cite{Ammar2016a}, but we do not include them for brevity, and because they show similar relative trends.}

\subsection{Isomorphism and Learning} \label{s:isomorphism-and-learning}

Non-isomorphic embedding spaces have been attributed largely to typological differences between languages \cite{sogaard2018limitations,Patra2019,Ormazabal2019}.
We hypothesise that non-isomorphism is not solely an intrinsic property of dissimilar languages, but also a result of a poorly conditioned training setup. In particular, languages that are regarded as being dissimilar to English, i.e. non-Indo-European languages, are often also low-resource languages where comparatively few samples for learning word embeddings are available.\footnote{There are obvious exceptions to this, such as Mandarin.} As a result, embeddings trained for low-resource languages may often not match the quality of their high-resource counterparts, and may thus constitute the main challenge when mapping embedding spaces. To investigate this hypothesis, we consider different aspects of poor conditioning as follows.

\vspace{1.8mm}
\noindent \textbf{Corpus size} It has become standard to align monolingual word embeddings trained on Wikipedia \cite{Glavas2019,Zhang2019}. As can be seen in Figure~\ref{fig:token-counts-vs-scores}, and also in Table~\ref{tab:wikis}, Wikipedias of low-resource languages are more than a magnitude smaller than Wikipedias of high-resource languages.\footnote{Recent initiatives replace training on Wikipedia with training on larger CommonCrawl data \cite{Grave:2018lrec,Conneau:2019arxiv}, but the large differences in corpora sizes between high-resource and low-resource languages are not removed.} Corpus size has been shown to play a role in the performance of monolingual embeddings \cite{Sahlgren2016}, but it is unclear how it influences their structure and isomorphism.

\vspace{1.8mm}
\noindent \textbf{Training duration} As it is generally too expensive to tune hyper-parameters separately for each language, monolingual embeddings are typically trained for the same number of epochs in large-scale studies. As a result, word embeddings of low-resource languages may be ``under-trained''.

\vspace{1.8mm}
\noindent \textbf{Preprocessing} Different forms of preprocessing have been shown to aid in learning a mapping \cite{Artetxe2018aaai,Vulic2019,Zhang2019}. Consequently, they may also influence the isomorphism of the vector spaces.

\vspace{1.8mm}
\noindent \textbf{Topical skew} The Wikipedias of low-resource languages may be dominated by few contributors, skewed towards particular topics, or generated automatically.\footnote{As one prominent example, a bot has generated most articles in the Swedish, Cebuano, and Waray Wikipedias.} Embeddings trained on different domains are known to be non-isomorphic \cite{sogaard2018limitations,Vulic2019}. A topical skew may thus also make embedding spaces harder to align.


\section{Simulating Low-resource Settings} 
\label{s:simulated-low-resource}

As low-resource languages---by definition---have only a limited amount of data available, we cannot easily control for all aspects using only a low-resource language. Instead, we modify the training setup of a high-resource language to simulate a low-resource scenario. For most of our experiments, we use English (\textsc{en}) as the source language and modify the training setup of Spanish (\textsc{es}). Additional results where we modify the training setup of English instead are available in the appendix; they further corroborate our key findings. We choose this language pair as both are similar, i.e. Indo-European, high-resource, and BLI performance is typically very high. Despite this high performance, unlike English, Spanish is a highly inflected language. In order to inspect if similar patterns also hold across typologically more dissimilar languages, we also conduct simulation experiments with two other target languages with large Wikipedias in lieu of Spanish: Japanese (\textsc{ja}, an agglutinative language) and Arabic (\textsc{ar}, introflexive). 

\begin{table}[t]
\def\arraystretch{0.99}
\centering
{\footnotesize
\begin{tabularx}{0.48\textwidth}{YY c}
\toprule
\multicolumn{2}{c}{\textsc{es} Wikipedia sample} & {} \\
\cmidrule(lr){1-2} 
{\# Sentences} & {\# Tokens} & {Comparable Wikis} \\
\cmidrule(lr){1-2} \cmidrule(lr){3-3} 
{50k} & {1.3M} & {Amharic, Yoruba, Khmer} \\
{100k} & {2.7M} & {Ilocano, Punjabi} \\
{200k} & {5.4M} & {Burmese, Nepali, Irish} \\
{500k} & {13.4M} & {Telugu, Tatar, Afrikaans} \\
{1M} & {26.8M} & {Armenian, Uzbek, Latvian} \\
{2M} & {53.7M} & {Croatian, Slovak, Malay} \\
{5M} & {134.1M} & {Finnish, Indonesian} \\
{10M} & {268.3M} & {Catalan, Ukrainian} \\

\bottomrule
\end{tabularx}
}
\vspace{-1mm}
\caption{Spanish Wikipedia samples of different sizes and comparable Wikipedias in other languages.}
\label{tab:wikis}
\vspace{-1mm}
\end{table}

When controlling for \textit{corpus size}, we subsample the target language (i.e., Spanish, Japanese, or Arabic) Wikipedia to obtain numbers of tokens comparable to low-resource languages as illustrated in Table~\ref{tab:wikis}. When controlling for \textit{training duration}, we take snapshots of the ``under-trained'' vector spaces after seeing an exact number of $M$ word tokens (i.e., after performing $M$ \textit{updates}). 

To control for \textit{topical skew}, we need to sample similar documents as in low-resource languages. To maximise topical overlap, we choose low-resource languages that are spoken in similar regions as Spanish and whose Wikipedias might thus also focus on similar topics---specifically Basque (\textsc{eu}), Galician (\textsc{gl}), and Quechua (\textsc{qu}). These four languages have very different morphology. Quechua is an agglutinative language, while Spanish, Galician, and Basque are highly inflected. Basque additionally employs case marking and derivation. If non-isomorphism was entirely explained by language dissimilarity, we would expect even low-resource versions of Spanish to have high BLI performance with English. We repeat the same experiment with another set of languages with distinct properties but spoken in similar regions: Bengali, Urdu, and Tamil.


\vspace{1.8mm}
\noindent \textbf{Typological differences} however, may still explain part of the difference in performance. For instance, as we cannot simulate Basque by changing the typological features of Spanish\footnote{\citet{Ravfogel2019} generate synthetic versions of English that differ from English in a single typological parameter. This process requires a treebank and is infeasible for all typological parameters of a language.}, we instead make Spanish, Basque, Galician, and Quechuan ``morphologically similar'': we remove inflections and case marking through lemmatisation. We follow the same process for Bengali, Urdu, and Tamil.


\section{Experiments and Analyses}
\label{s:experimental}
\begin{figure*}[!t]
    \centering
    \begin{subfigure}[!t]{0.328\linewidth}
        \centering
        \includegraphics[width=0.98\linewidth]{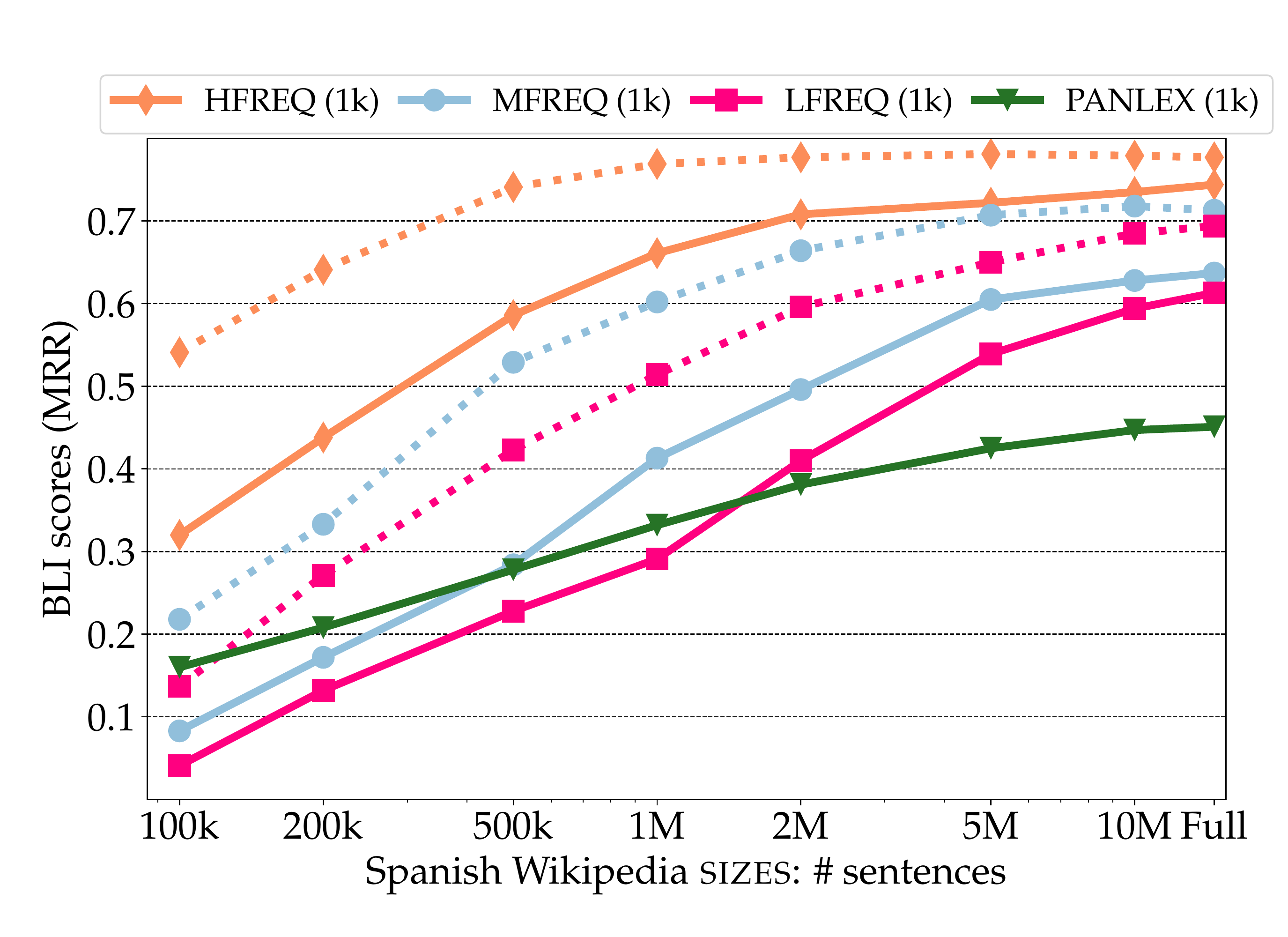}
        \caption{EN $\rightarrow$ ES (1k)}
        \label{fig:es-sizes-1k}
    \end{subfigure}
    \begin{subfigure}[!t]{0.328\textwidth}
        \centering
        \includegraphics[width=0.98\linewidth]{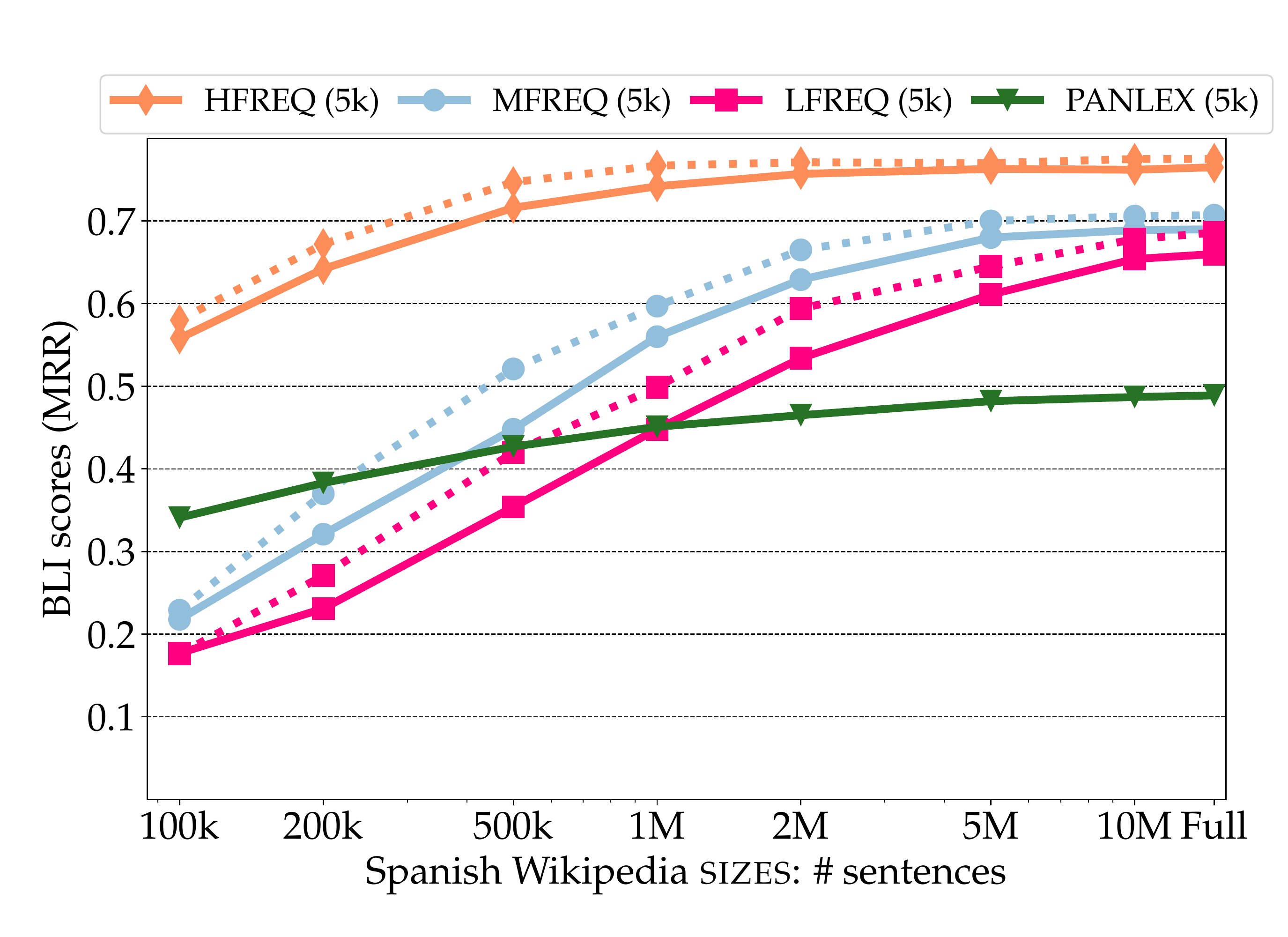}
        \caption{EN $\rightarrow$ ES (5k)}
        \label{fig:es-sizes-5k}
    \end{subfigure}
        \begin{subfigure}[!t]{0.328\textwidth}
        \centering
        \includegraphics[width=0.98\linewidth]{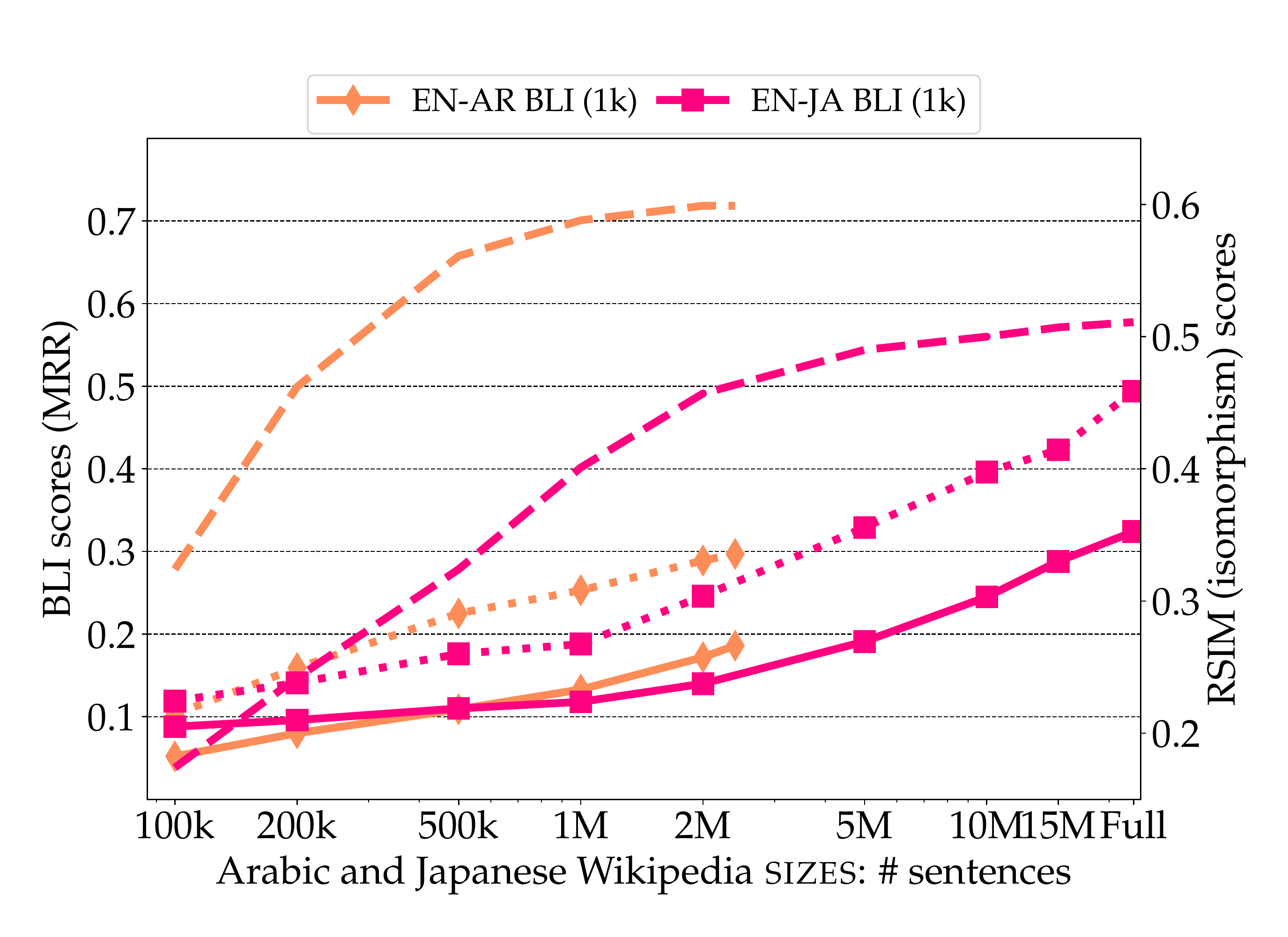}
        \caption{EN $\rightarrow$ AR/JA (1k)}
        \label{fig:arja-sizes-1k}
    \end{subfigure}
    \vspace{-1mm}
    \caption{Impact of \textbf{dataset size} on BLI when aligning \textsc{es}, \textsc{ar}, and \textsc{ja} vector spaces fully trained on corpora of different sizes (obtained through sampling from the full corpus) to an \textsc{en} space fully trained on complete data. We report scores without self-learning (solid lines) and with self-learning (dotted lines; same colour) with seed dictionary sizes of \textbf{(a)} 1k and \textbf{(b)} 5k on our \textsc{en}--\textsc{es} BLI evaluation sets, while the corresponding isomorphism scores are provided in Figure~\ref{fig:sizes-enes-iso} for clarity. \textbf{(c)} We again report scores without and with self-learning on \textsc{en}--\textsc{ar/ja} BLI evaluation sets from the MUSE benchmark with 1k seed translation pairs. The results with 5k seed pairs for \textsc{en}--\textsc{ar/ja} are available in the appendix. Dashed lines without any marks show isomorphism scores (computed by RSIM; higher is better) computed across different \textsc{ar} and \textsc{ja} snapshots.}
    \vspace{-0.5mm}
\label{fig:dataset-size}
\end{figure*}
\begin{figure*}[!t]
    \centering
    \begin{subfigure}[!t]{0.328\linewidth}
        \centering
        \includegraphics[width=0.98\linewidth]{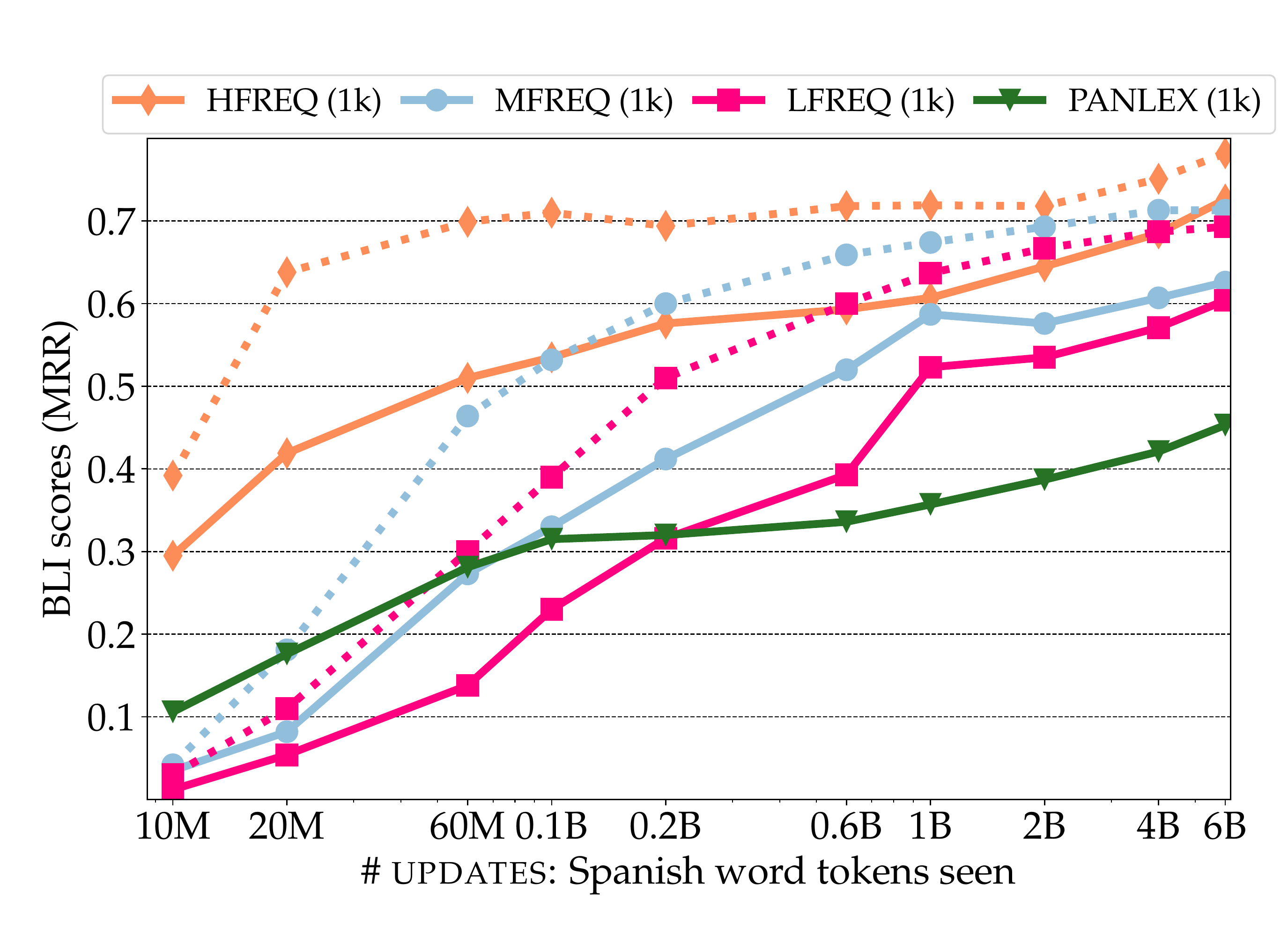}
        \caption{EN $\rightarrow$ ES (1k)}
        \label{fig:esupdates-1k}
    \end{subfigure}
    \begin{subfigure}[!t]{0.328\textwidth}
        \centering
        \includegraphics[width=0.98\linewidth]{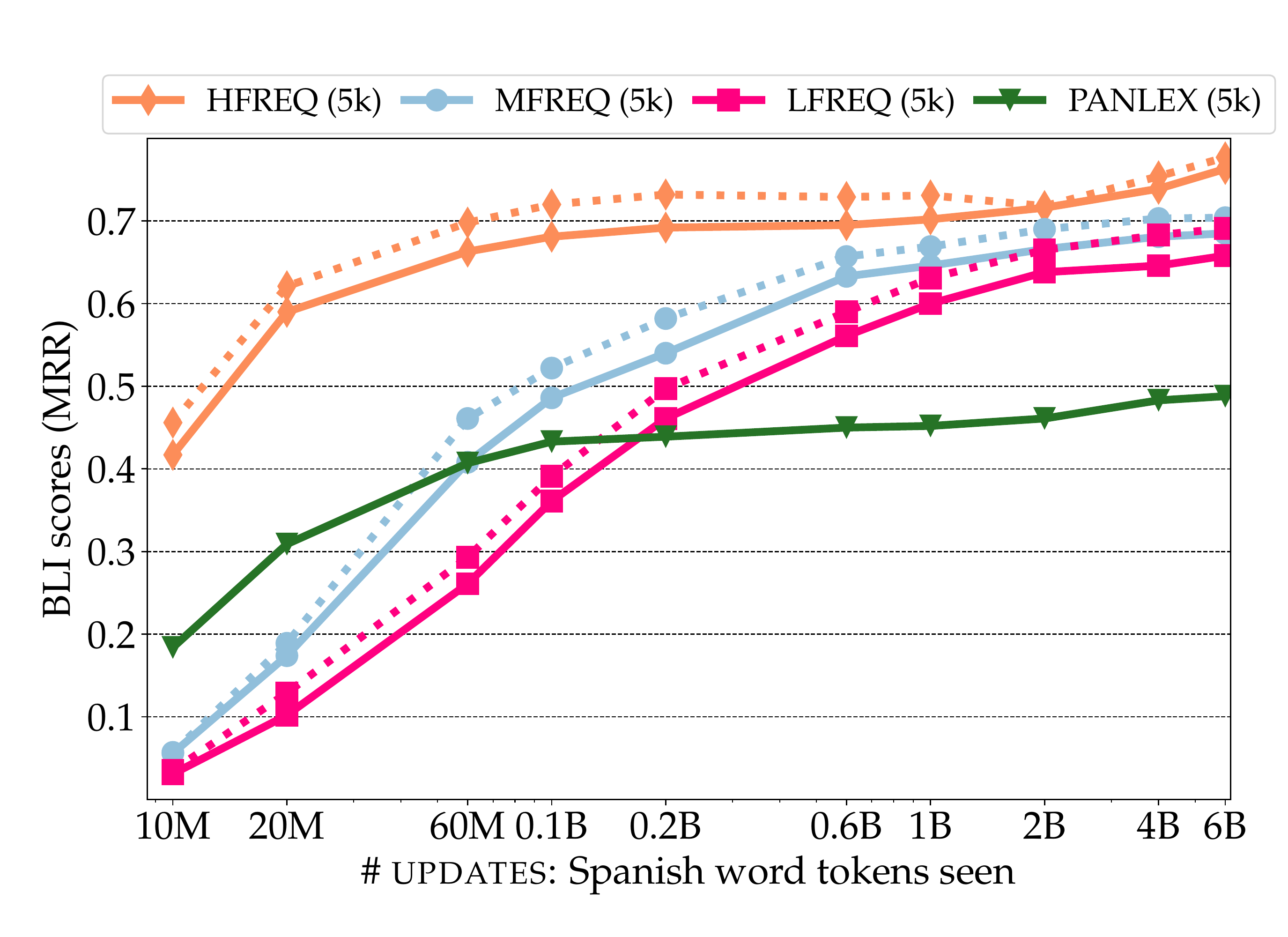}
        \caption{EN $\rightarrow$ ES (5k)}
        \label{fig:esupdates-5k}
    \end{subfigure}
    \begin{subfigure}[!t]{0.328\textwidth}
        \centering
        \includegraphics[width=0.98\linewidth]{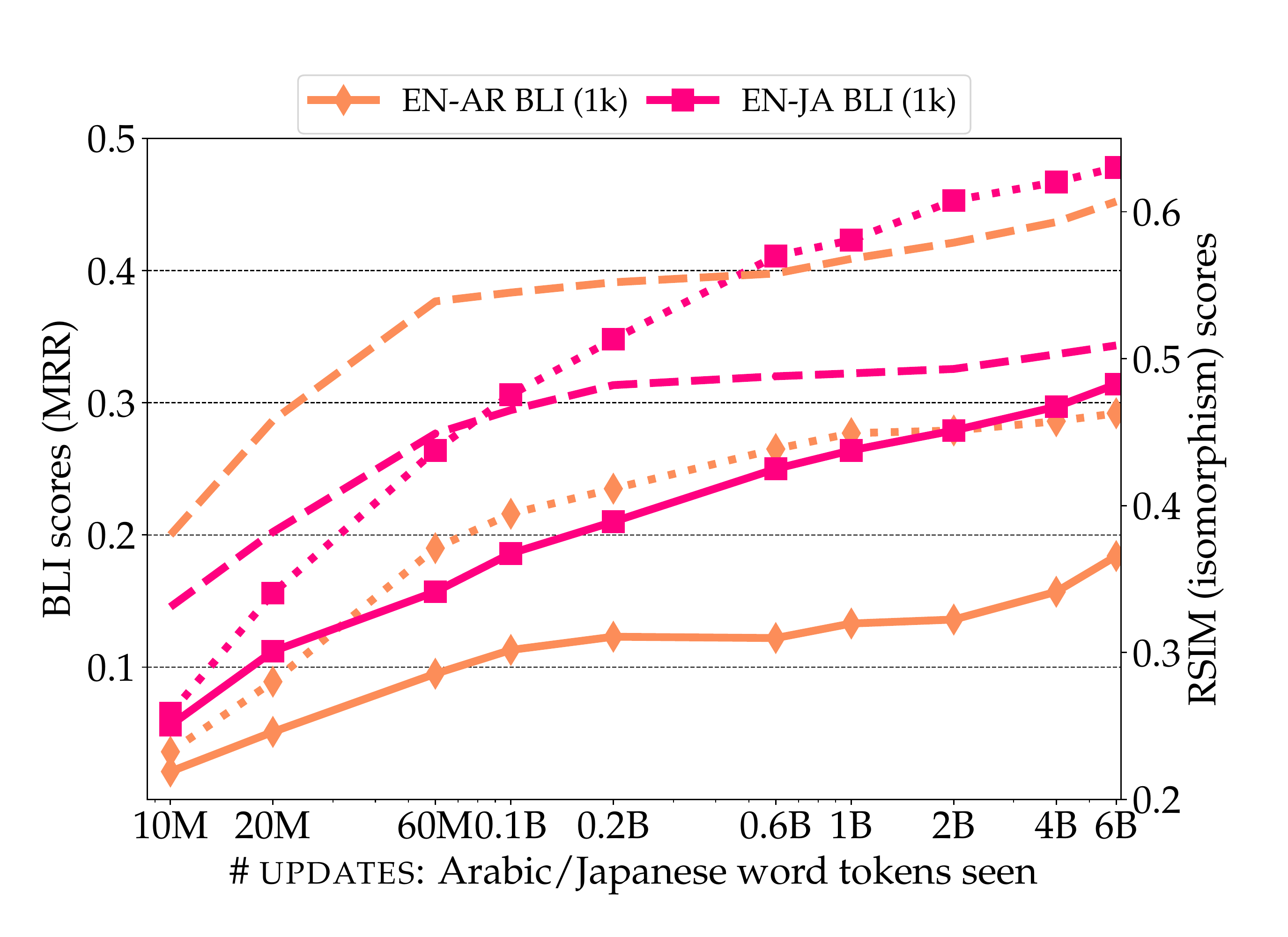}
        \caption{EN $\rightarrow$ AR/JA (1k)}
        \label{fig:arja-updates-1k}
    \end{subfigure}
    \vspace{-1mm}
    \caption{Impact of \textbf{training duration} on BLI when aligning a partially trained Spanish (\textsc{es}), Arabic (\textsc{ar}), and Japanese (\textsc{ja}) vector space, where snapshots are taken after seeing $M$ word tokens in training, to the fully trained \textsc{en} space. We report scores without self-learning (solid lines) and with self-learning (dotted lines; same colour) with seed dictionary sizes of \textbf{(a)} 1k and \textbf{(b)} 5k on our \textsc{en}--\textsc{es} BLI evaluation sets. For clarity, the corresponding isomorphism scores (and impact of training duration on isomorphism of vector spaces) over the same training snapshots for Spanish are shown in Figure~\ref{fig:preproc}. \textbf{(c)} We again report scores without and with self-learning on \textsc{en}--\textsc{ar/ja} BLI evaluation sets from the MUSE benchmark with 1k seed translation pairs. The results with 5k seed pairs for \textsc{en}--\textsc{ar/ja} are available in the appendix. Dashed lines without any marks show isomorphism scores (computed by RSIM; higher is better) computed across different \textsc{ar} and \textsc{ja} snapshots.}
    \vspace{-0.5mm}
\label{fig:training-duration}
\end{figure*}

\subsection{Experimental Setup}

\noindent \textbf{Embedding algorithm} Previous work has shown that learning embedding spaces with different hyper-parameters leads to non-isomorphic spaces \cite{sogaard2018limitations,Hartmann2018}. To control for this aspect, we train monolingual embeddings with fastText in the standard setup (skip-gram, character n-grams of sizes 3 to 6, a learning rate of 0.025, 15 negative samples, a window size of 5) \cite{Bojanowski:2017tacl}.\footnote{We ran experiments and observed similar results with \texttt{word2vec} algorithms \cite{Mikolov:2013nips}, GloVe \cite{Pennington:2014emnlp} and fastText CBOW \cite{Grave:2018lrec}.} Unless specified otherwise, we train for 15 epochs.

\vspace{1.8mm}
\noindent \textbf{Mapping algorithm} We use the supervised variant of VecMap \cite{Artetxe2018} for our experiments, which is a robust and competitive choice according to the recent empirical comparative studies \cite{Glavas2019,Vulic2019,Hartmann:2019neurips}. VecMap learns an orthogonal transformation based on a seed translation dictionary with additional preprocessing and postprocessing steps, and it can additionally enable \textit{self-learning} in multiple iterations. For further details we refer the reader to the original work \cite{Artetxe2018}.

\vspace{1.8mm}
\noindent \textbf{Evaluation} We measure isomorphism between monolingual spaces using the previously described intrinsic measures: eigenvector similarity (EVS), Gromov-Hausdorff distance (GH), and relational similarity (RSIM). In addition, we evaluate on bilingual lexicon induction (BLI), a standard task for evaluating cross-lingual word representations. Given a list of $N_s$ source words, the task is to find the corresponding translation in the target language as a nearest neighbour in the cross-lingual embedding space. The list of retrieved translations is then compared against a gold standard dictionary. Following prior work \cite{Glavas2019}, we employ mean reciprocal rank (MRR) as evaluation measure, and use cosine similarity as similarity measure.



\vspace{1.8mm}
\noindent \textbf{Training and test dictionaries} Standard BLI test dictionaries over-emphasise frequent words \cite{Czarnowska2019,Kementchedjhieva2019} whose neighbourhoods may be more isomorphic \cite{Nakashole2018}. We thus create new evaluation dictionaries for English--Spanish that consist of words in different frequency bins: we sample \en words for 300 translation pairs respectively from (i) the top 5k words of the full English Wikipedia (\hf); (ii)  the interval $[$10k, 20k$]$ (\mf); (iii) the interval $[$20k, 50k$]$ (\lf).
The entire dataset (\af; 900 pairs) consists of (i) + (ii) + (iii). We exclude named entities as they are over-represented in many test sets \cite{Kementchedjhieva2019} and include nouns, verbs, adjectives, and adverbs in all three sets. All 900 words have been carefully manually translated and double-checked by a native Spanish speaker. There are no duplicates.
We also report BLI results on the PanLex test lexicons \cite{Vulic2019}. 

For English--Spanish, we create training dictionaries of sizes 1k and 5k based on PanLex \cite{Kamholz2014panlex} following \citet{Vulic2019}. We exclude all words from $\af$ from the training set. For \textsc{en}--\textsc{ja/ar} BLI experiments, we rely on the standard training and test dictionaries from the MUSE benchmark \cite{Conneau2018}. Isomorphism scores with RSIM for \textsc{en}--\textsc{ja/ar} are computed on a fixed random sample of 1k one-to-one translations from the respective MUSE training dictionary.\footnote{Note that the \textit{absolute} isomorphism scores between different pairs of languages are not directly comparable. The focus of the experiments is to follow the patterns of isomorphism change for each language pair separately.} For learning monolingual embeddings, we use tokenised and sentence-split Polyglot Wikipedias \cite{polyglot:2013:ACL-CoNLL}. In \textsection \ref{sec:topical_skew}, we process Wiki dumps, using Moses for tokenisation and sentence splitting. For lemmatisation of Spanish, Basque, Galician, Tamil, and Urdu we employ the UDPipe models \cite{straka2017tokenizing}. For Quechua and Bengali, we utilise the unsupervised Morfessor model provided by Polyglot NLP.


\subsection{Impact of Corpus Size} 
To evaluate the impact of the corpus size on vector space isomorphism and BLI performance, we shuffle the target language (i.e., Spanish, Arabic, Japanese) Wikipedias and take $N$ sentences where $N \in \{$10k, 20k, 500k, 100k, 500k, 1M, 2M, 10M, 15M$\}$ corresponding to a range of low-resource languages (see Table \ref{tab:wikis}). Each smaller dataset is a subset of the larger one. We learn target language embeddings for each sample, map them to the English embeddings using dictionaries of sizes 1k and 5k and supervised VecMap with and without self-learning, and report their BLI performance and isomorphism scores in Figure~\ref{fig:dataset-size}. Both isomorphism and BLI scores improve with larger training resources.\footnote{We observe similar trends when estimating isomorphism on the 5k and 10k most frequent words in both languages.} Performance is higher with a larger training dictionary and self-learning but shows a similar convergence behaviour irrespective of these choices. What is more, despite different absolute scores, we observe a similar behaviour for all three language pairs, demonstrating that our intuition holds across typologically diverse languages. 

In English--Spanish experiments performance on frequent words converges relatively early, between 1-2M sentences, while performance on medium and low-frequency words continues to increase with more training data and only plateaus around 10M sentences. Self-learning improves BLI scores, especially in low-data regimes. 
Note that isomorphism scores increase even as BLI scores saturate, which we discuss in more detail in \S\ref{s:discussion}.


\begin{figure*}[!t]
    \centering
    \begin{subfigure}[!t]{0.458\linewidth}
        \centering
        \includegraphics[width=0.98\linewidth]{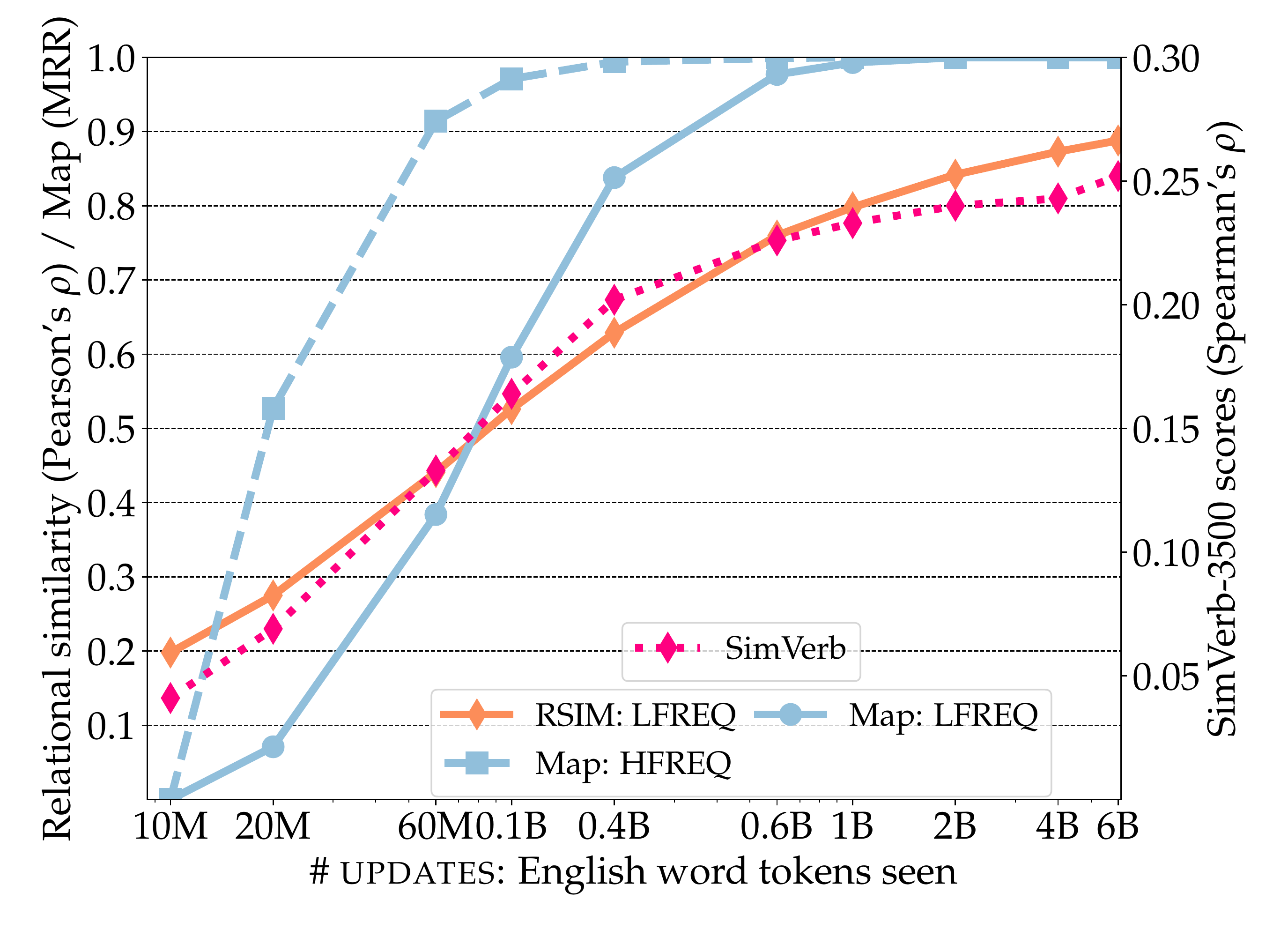}
        \caption{Updates: \textsc{en}}
        \label{fig:en-monoupdates}
    \end{subfigure}
    \begin{subfigure}[!t]{0.458\textwidth}
        \centering
        \includegraphics[width=0.98\linewidth]{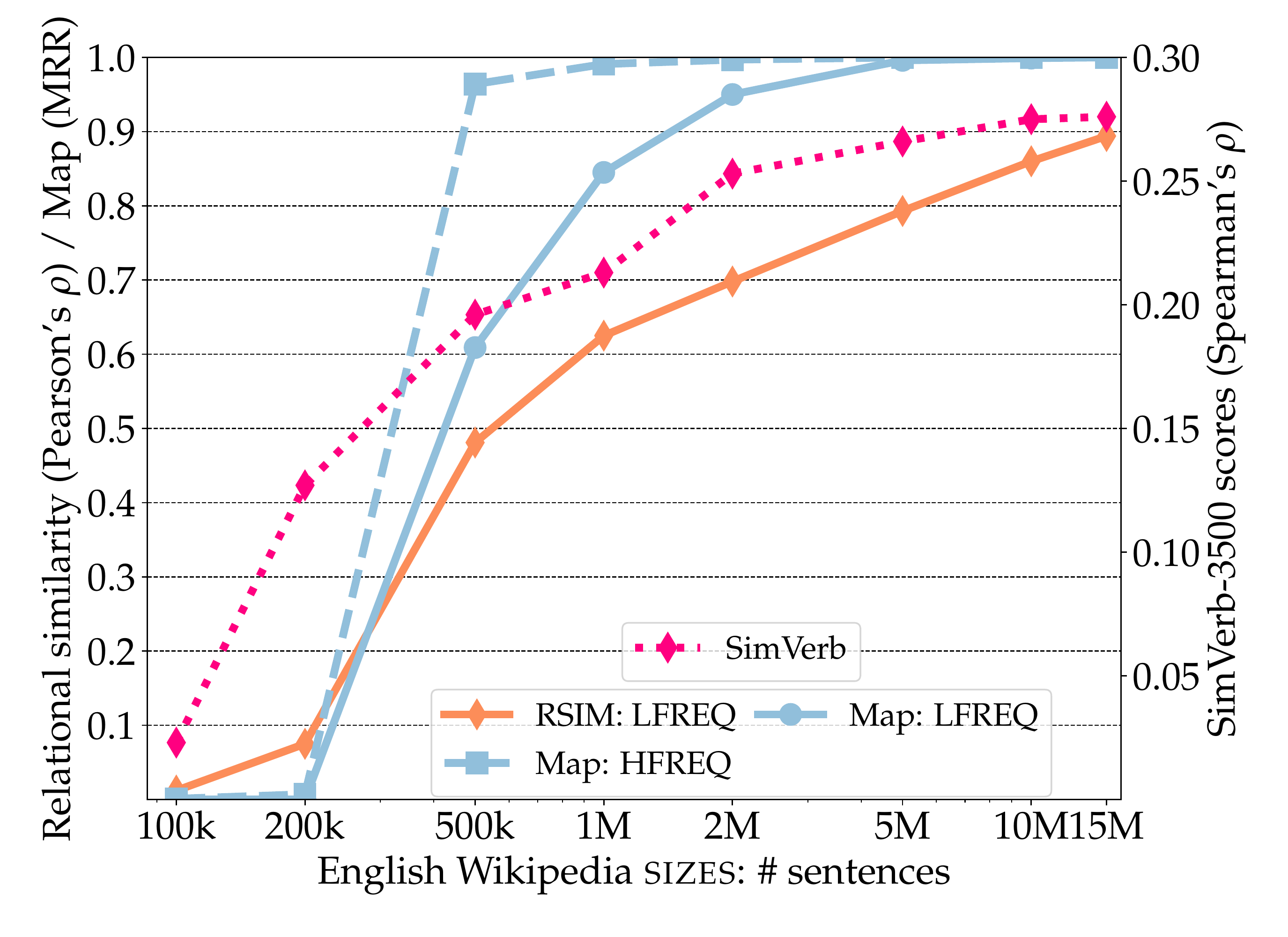}
        \caption{Different Wiki sizes: \textsc{en}}
        \label{fig:en-monosizes}
    \end{subfigure}
    \vspace{-1mm}
    \caption{Monolingual ``control'' experiments when aligning \textbf{(a)} a partially trained \textsc{en} vector space (after $M$ updates, that is, seen word tokens) to a fully trained vector space, and \textbf{(b)} an \textsc{en} vector space fully trained on Wikipedia of different sizes (number of sentences). We show RSIM scores, mapping performance (i.e., monolingual ``BLI'') on \hf and \lf \textsc{en} words, and monolingual word similarity scores on SimVerb-3500.}
    \vspace{-1mm}
\label{fig:monolingual}
\end{figure*}

\subsection{Impact of Training Duration} To analyse the effect of under-training, we align English embeddings with target language embeddings that were trained for a certain number of iterations/updates and compute their BLI scores. The results for the three language pairs are provided in Figure~\ref{fig:training-duration}. As monolingual vectors are trained for longer periods, BLI and isomorphism scores improve monotonously, and this holds for all three language pairs. Even after training for a large number of updates, BLI and isomorphism scores do not show clear signs of convergence. Self-learning again seems beneficial for BLI, especially at earlier, ``under-training'' stages.

\begin{figure*}[!t]
    \centering
    \begin{subfigure}[!t]{0.318\linewidth}
        \centering
        \includegraphics[width=0.98\linewidth]{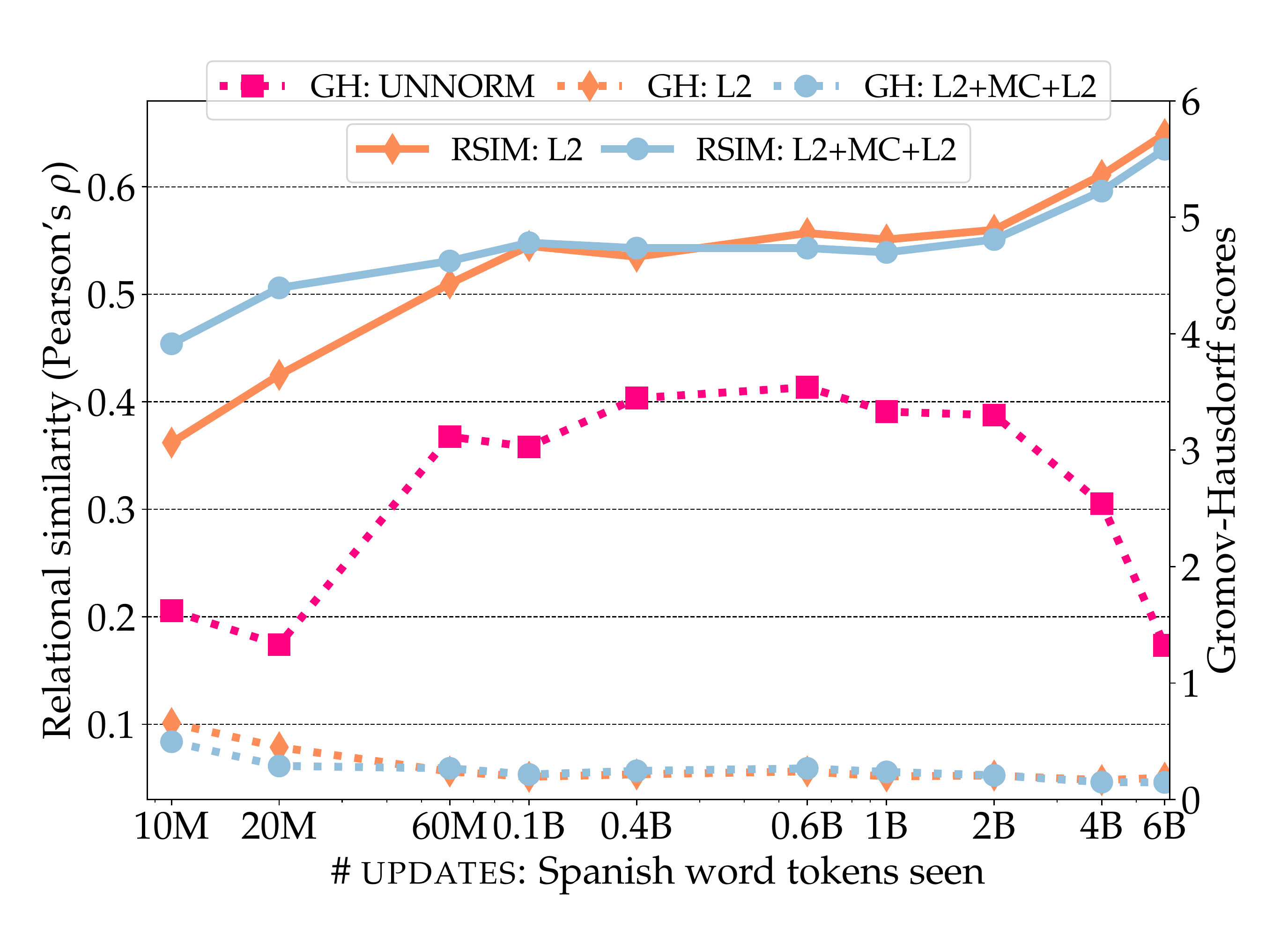}
        \caption{\hf subset (\es)}
        \label{fig:iso-es-hfreq}
    \end{subfigure}
    \begin{subfigure}[!t]{0.318\textwidth}
        \centering
        \includegraphics[width=0.98\linewidth]{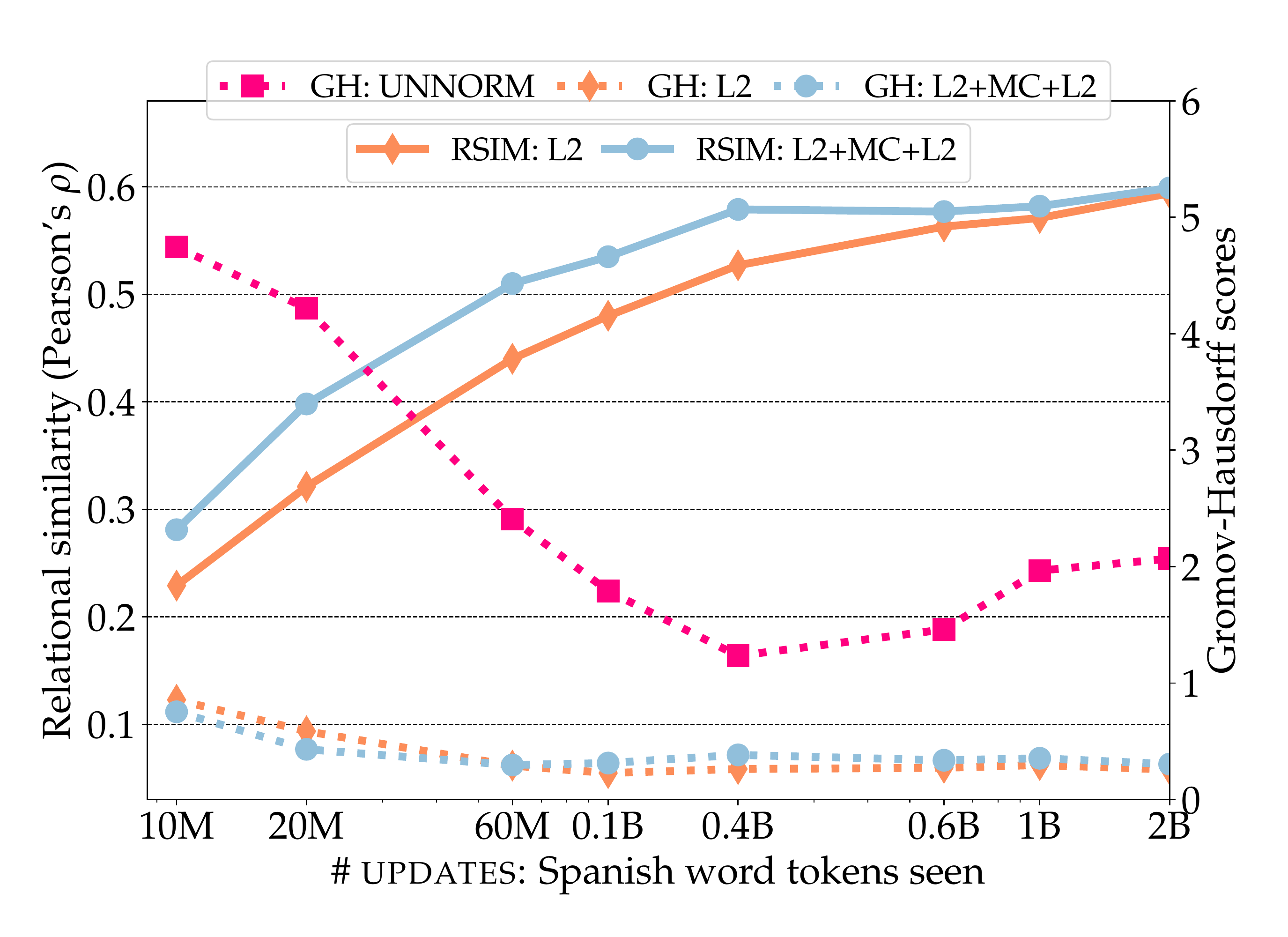}
        \caption{\mf subset (\es)}
        \label{fig:iso-es-mfreq}
    \end{subfigure}
        \begin{subfigure}[!t]{0.318\textwidth}
        \centering
        \includegraphics[width=0.98\linewidth]{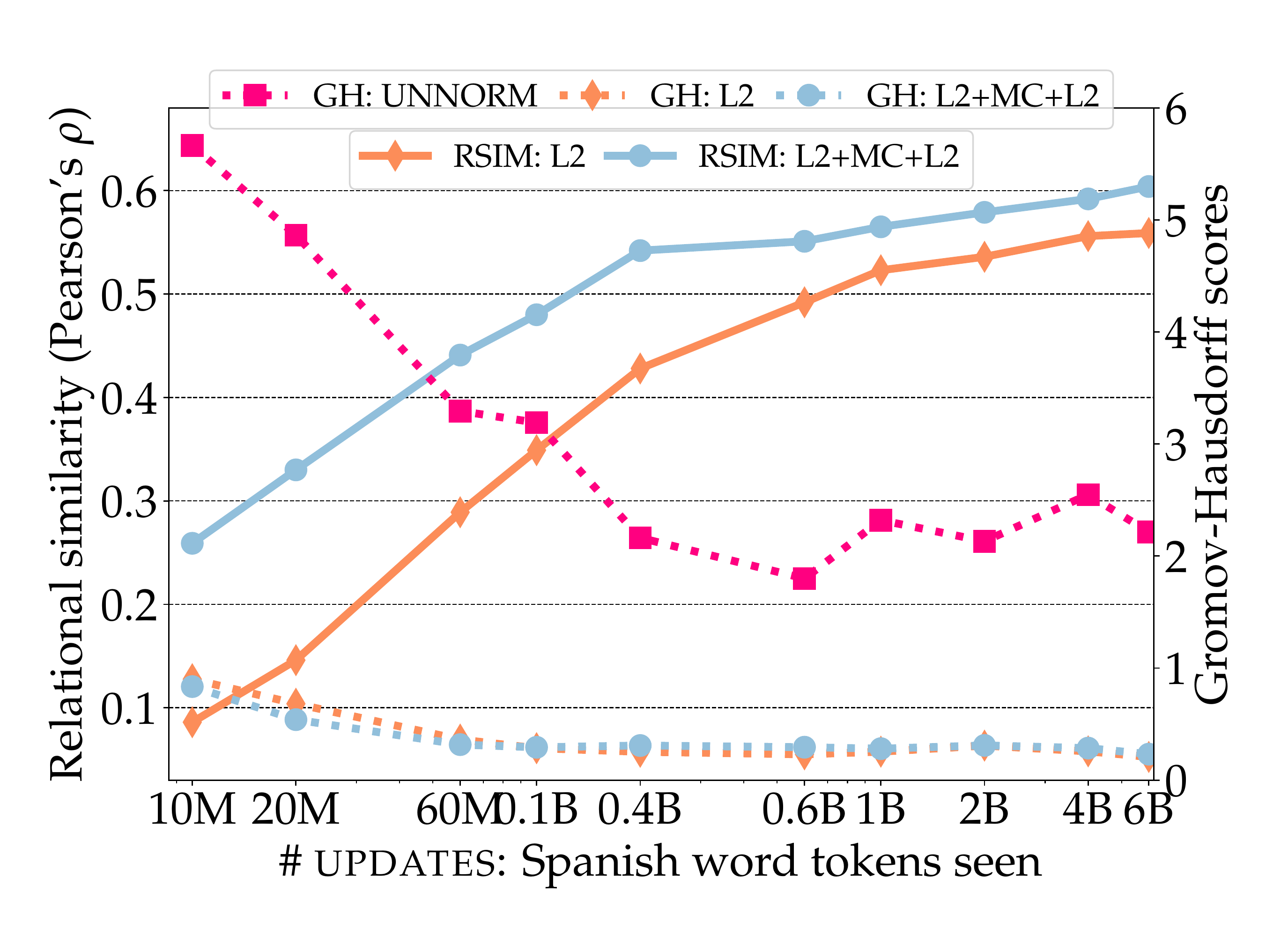}
        \caption{\lf subset (\es)}
        \label{fig:iso-es-lfreq}
    \end{subfigure}
    \vspace{-0.5mm}
    \caption{Impact of different monolingual vector space preprocessing strategies on isomorphism scores when aligning a partially trained \textsc{es} vector space, where snapshots are taken after seeing $M$ word tokens in training, to a fully trained \textsc{en} vector space. We report RSIM (solid; higher is better, i.e., more isomorphic) and GH distance (dashed; lower is better) on \textbf{(a)} \hf, \textbf{(b)} \mf, and \textbf{(c)} \lf test sets.
    }
    \vspace{-1mm}
\label{fig:preproc}
\end{figure*}

\begin{figure}[!t]
    \centering
    \includegraphics[width=0.96\linewidth]{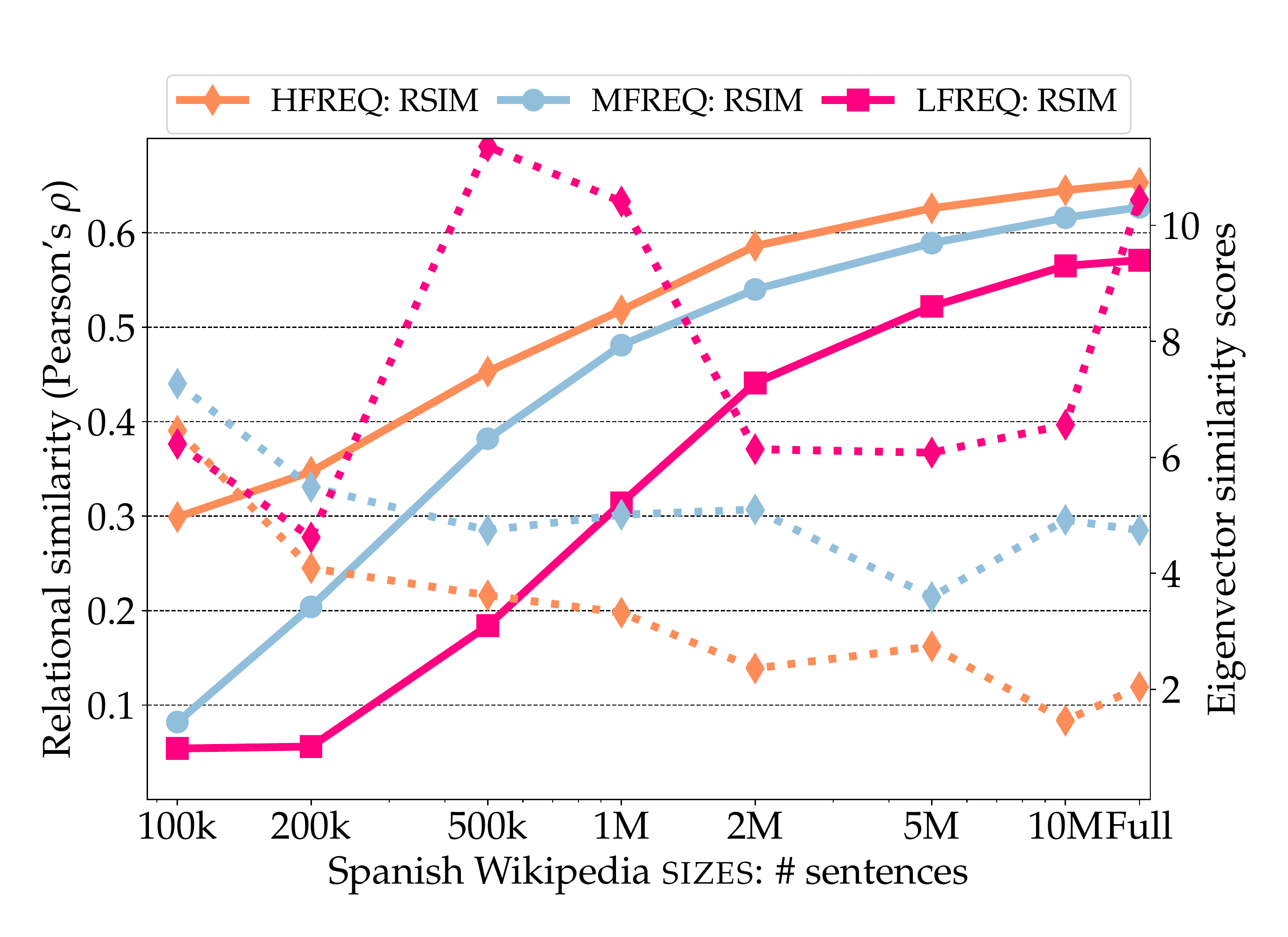}
    \vspace{-0.5mm}
    \caption{Impact of \textbf{dataset size} on vector space isomorphism when aligning an \textsc{es} vector space fully trained on corpora of different sizes to an \textsc{en} space fully trained on complete data. RSIM (solid lines; higher is better, i.e. more isomorphic) and Eigenvector similarity (dotted lines; lower is better) scores are reported. See Figures~\ref{fig:es-sizes-1k}-\ref{fig:es-sizes-5k} for the corresponding \textsc{en--es} BLI scores.}
    \vspace{-0.5mm}
    \label{fig:sizes-enes-iso}
\end{figure}

\subsection{Impact on Monolingual Mapping} 
As a control experiment, we repeat the two previous experiments controlling for corpus size and training duration when mapping an English embedding space to another \textsc{en} embedding space. Previous work \cite{Hartmann2018} has shown that \textsc{en} embeddings learned with the same algorithm achieve a perfect monolingual ``BLI'' score of 1.0 (mapping \textsc{en} words to the same \textsc{en} word). If typological differences were the only factor affecting the structure of embedding spaces, we would thus expect to achieve a perfect score also for shorter training and smaller corpus sizes. For comparison, we also provide scores on a standard monolingual word similarity benchmark, SimVerb-3500 \cite{Gerz:2016emnlp}. We show results in Figure \ref{fig:monolingual}. We observe that BLI scores only reach 1.0 after 0.4B and 0.6B updates for frequent and infrequent words or with corpus sizes of 1M and 5M sentences respectively, which is more than the size of most low-resource language Wikipedias (Table \ref{tab:wikis}). This clearly shows that even aligning \textsc{en} to \textsc{en} is challenging in a low-resource setting due to different vector space structures, and we cannot attribute performance differences to typological differences in this case. 

\begin{table*}[t]
\def\arraystretch{0.99}
{\fontsize{7.9pt}{7.9pt}\selectfont
\begin{tabular}{l l ccc l ccc l ccc}
\toprule
 && \multicolumn{3}{c}{Basque (\eu)} && \multicolumn{3}{c}{Quechua (\qu)} && \multicolumn{3}{c}{Galician (\gl)} \\
 && \en-\es & \en-\eu & \es/\eu gap & & \en-\es & \en-\qu & \es/\qu gap & & \en-\es & \en-\gl & \es/\gl gap \\
\midrule
Full wiki && 0.757 & 0.466 & 0.291 && 0.572 & 0.066 & 0.506 && 0.757  & 0.689 & 0.068 \\
Random sample && 0.662 & 0.411 & 0.251 && 0.037 & 0.054 & -0.017 && 0.680 & 0.663 & 0.017 \\
Comparable sample && 0.663 & 0.420 & 0.243 && 0.081 & 0.060 & 0.021 && 0.669 & 0.671 & -0.002 \\
Comp. sample + lemma && 0.533 & 0.404 & 0.129 &&  0.052 & 0.041 & 0.011 && 0.619 & 0.596 & 0.023 \\
\bottomrule
\end{tabular}
\begin{tabularx}{\linewidth}{l l YYY l YYY}
\toprule
 && \multicolumn{3}{c}{Tamil (\ta)} && \multicolumn{3}{c}{Urdu (\ur)} \\
 && \en-\bn & \en-\ta & \bn/\ta gap & & \en-\bn & \en-\ur & \bn/\ur gap \\
\midrule
Full wiki && {0.253} & {0.152} & {0.101} && {0.118} & {0.132} & {-0.014}  \\
Random sample && {0.193} & {0.124} & {0.069} && {0.076} & {0.093} & {-0.017}  \\
Comparable sample && {0.196} & {0.131} & {0.065} && {0.108} & {0.112} & {-0.004}  \\
Comp. sample + lemma && {0.152} & {0.121} & {0.031} && {0.072} & {0.070} & {0.002}  \\
\bottomrule
\end{tabularx}%
}
\vspace{-0.5mm}
\caption{BLI scores (MRR) when mapping from a fully trained \en embedding space to one trained on full Wikipedia corpora, random samples, and topic-adjusted comparable samples of the same size with and without lemmatisation for Spanish (\es) and Basque (\eu), Quechua (\qu), and Galician (\gl), respectively (\textbf{Top} table); Bengali (\bn) and Tamil (\ta), and \bn and Urdu (\ur) (\textbf{Bottom} table).}
\label{tab:topic-lemmas}
\vspace{-0.5mm}
\end{table*}
\begin{figure*}[t]
    \centering
    \begin{subfigure}[t]{0.318\linewidth}
        \centering
        \includegraphics[width=0.98\linewidth]{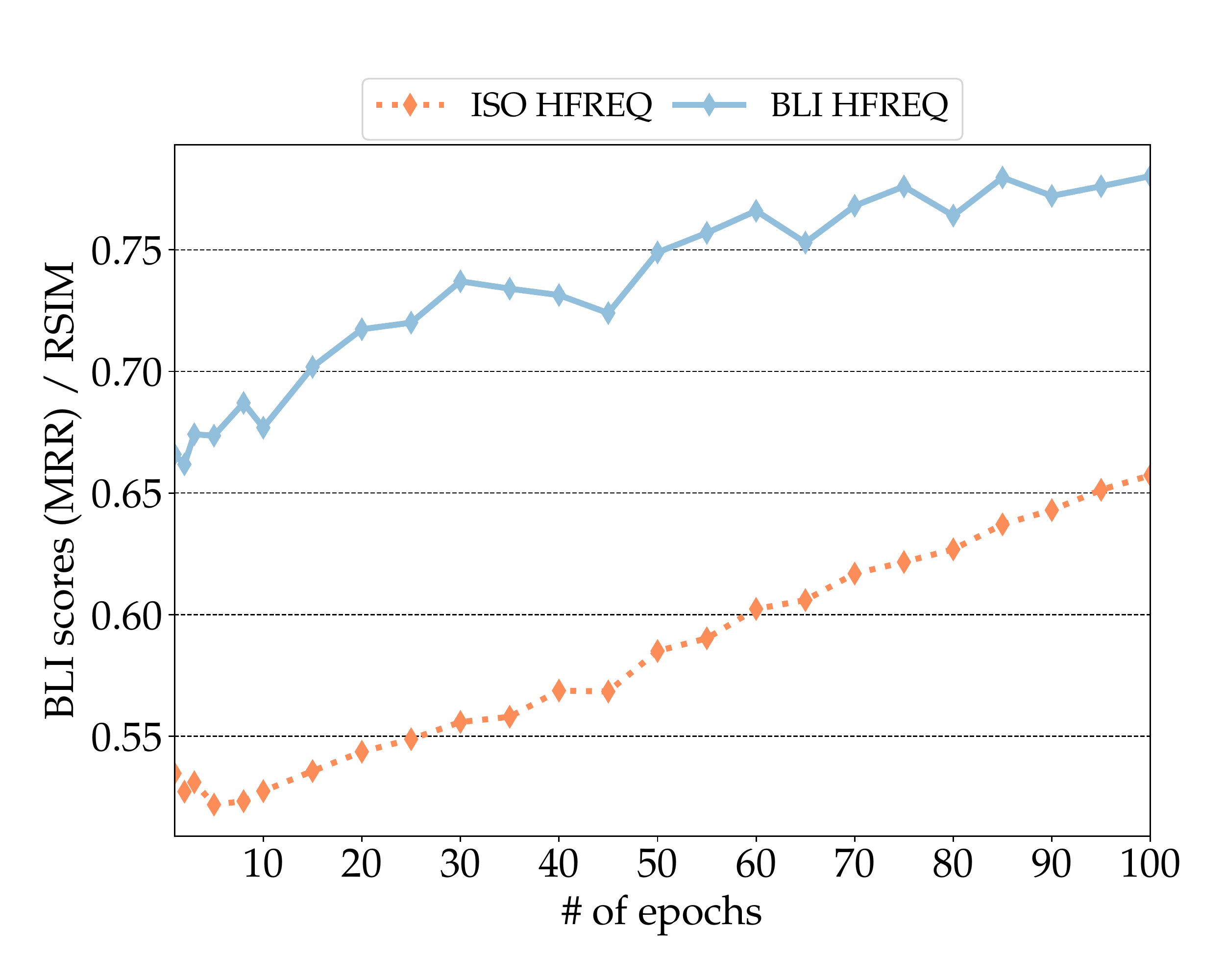}
        \caption{\hf subset (\es)}
        \label{fig:iso-es-hfreq}
    \end{subfigure}
    \begin{subfigure}[t]{0.318\textwidth}
        \centering
        \includegraphics[width=0.98\linewidth]{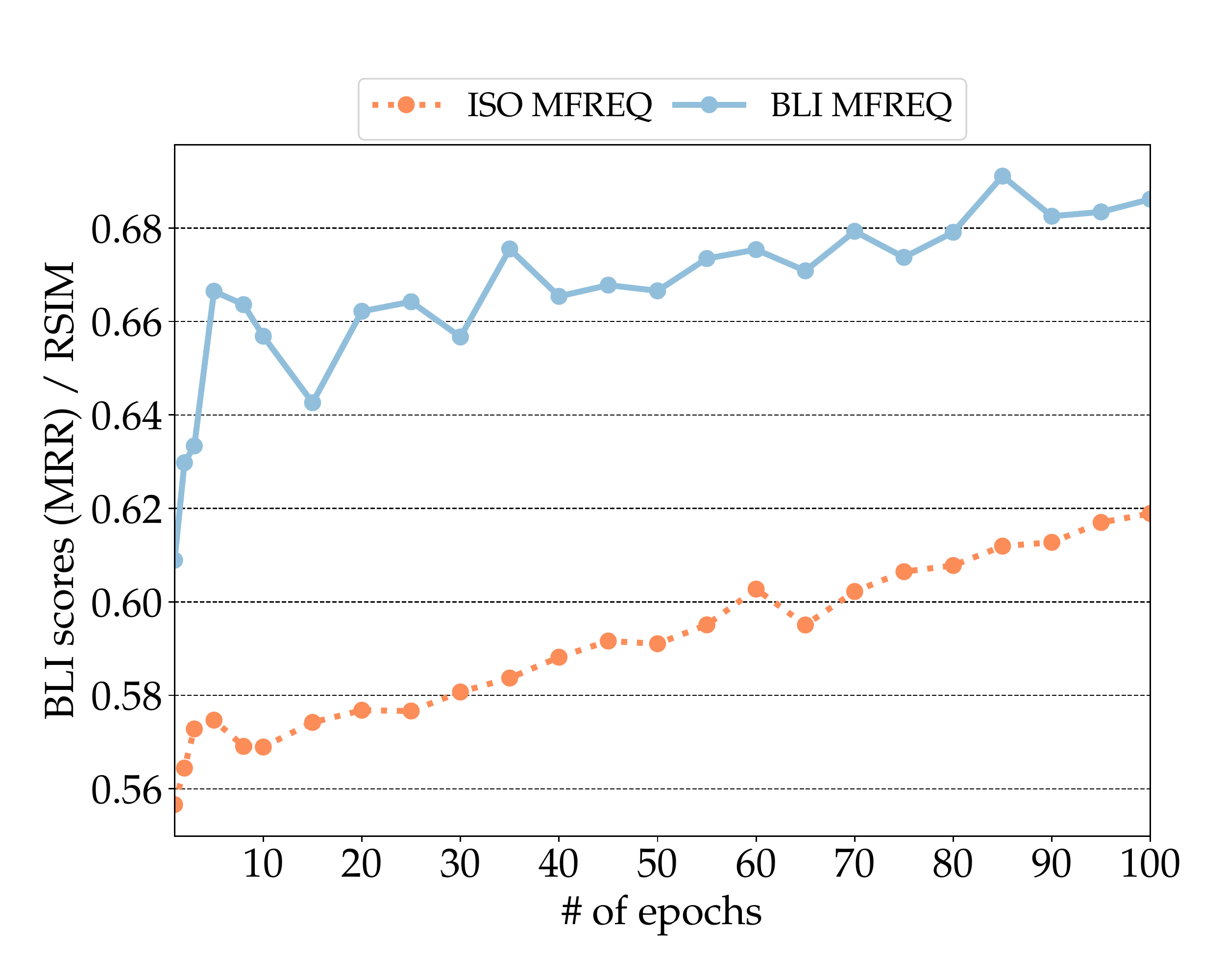}
        \caption{\mf subset (\es)}
        \label{fig:iso-es-mfreq}
    \end{subfigure}
        \begin{subfigure}[t]{0.318\textwidth}
        \centering
        \includegraphics[width=0.98\linewidth]{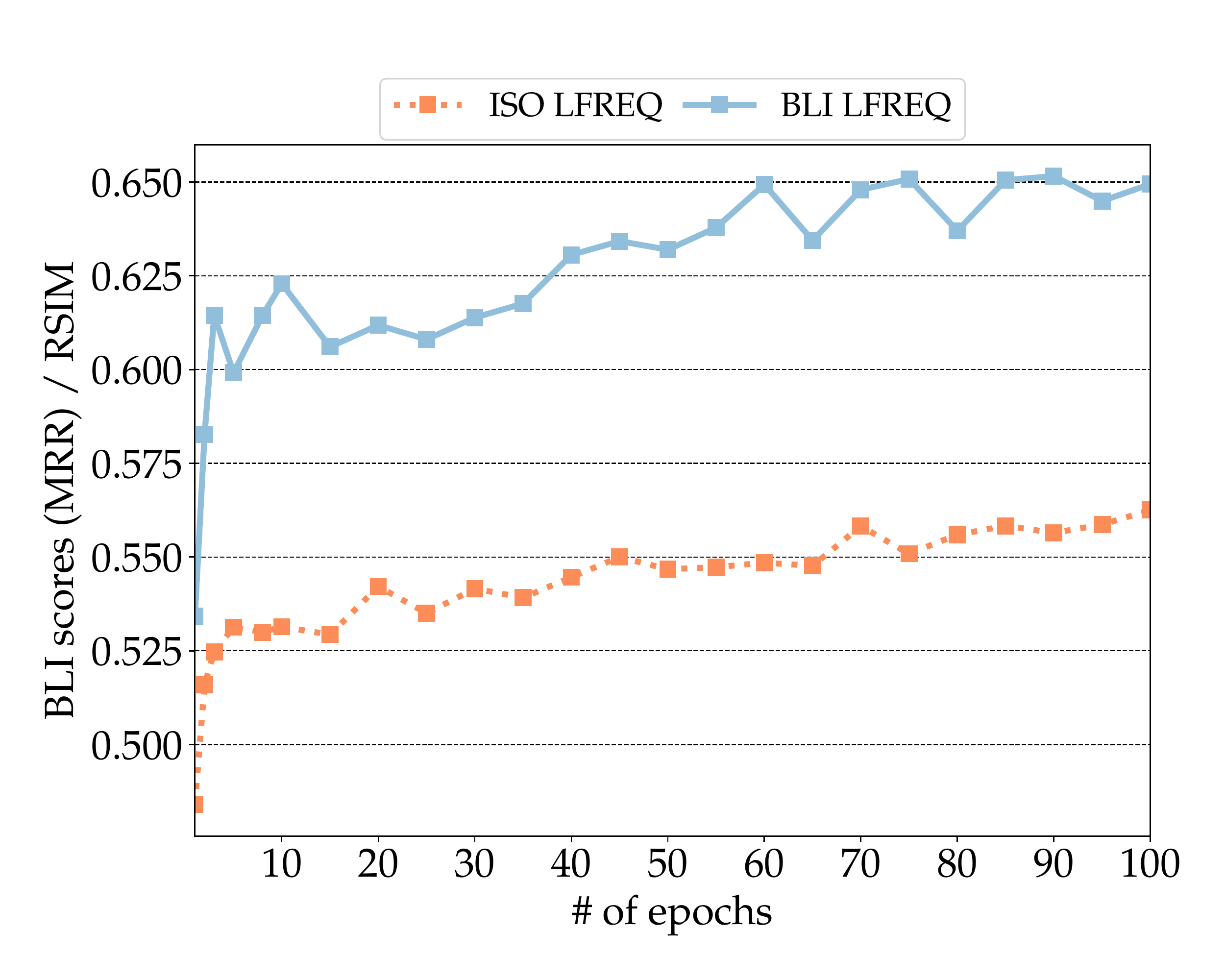}
        \caption{\lf subset (\es)}
        \label{fig:iso-es-lfreq}
    \end{subfigure}
    \vspace{-0.5mm}
    \caption{Monolingual learning dynamics and isomorphism (RSIM): We align a partially trained \textsc{es} vector space, after seeing $M$ word tokens, with a fully trained \textsc{en} vector space, and evaluate on (a) \hf, (b) \mf, and (c) \lf test sets. While BLI performance plateaus, the isomorphism score (computed with RSIM) does not.}
    \vspace{-1mm}
\label{fig:IB}
\end{figure*}

\subsection{Impact of Preprocessing} We next evaluate the impact of different forms of preprocessing. Specifically, we consider: \textbf{1)} No preprocessing (unnormalised vectors); \textbf{2)} Length normalisation (\elltwo) only (required for orthogonal Procrustes); \textbf{3)} \elltwo, mean centering (\mc), followed by \elltwo; used by VecMap \cite{Artetxe2018}; and \textbf{4)} Iterative normalisation \cite{Zhang2019}.

Iterative normalisation consists of multiple steps of \elltwo+\mc+\elltwo.\footnote{ \url{github.com/zhangmozhi/iternorm}} We have found it to achieve performance nearly identical to \elltwo+\mc+\elltwo, so we do not report it separately. We show results for the remaining methods in Figure~\ref{fig:preproc} and Figure~\ref{fig:sizes-enes-iso}. For GH, using no preprocessing leads to much less isomorphic spaces, particularly for infrequent words during very early training. For RSIM with cosine similarity, \elltwo is equivalent to no normalisation as cosine applies length normalisation. \elltwo+\mc+\elltwo leads to slightly better isomorphism scores overall compared to \elltwo alone, though it has a slightly negative impact on Gromov-Hausdorff scores over longer training duration. Most importantly, the results demonstrate that such preprocessing steps do have a profound impact on near-isomorphism between monolingual vector spaces.

\subsection{Impact of Topical Skew and Morphology} \label{sec:topical_skew}

To control for topical skew, we sample the Spanish Wikipedia so that its topical distribution is as close as possible to that of low-resource languages spoken in similar regions---Basque, Galician, and Quechua. To this end, for each language pair, we first obtain document-level alignments using the Wiki API\footnote{\url{https://www.wikidata.org/w/api.php}}. We only consider documents that occur in both languages. We then sample sentences from the \textsc{es} Wikipedia so that the number of tokens per document and the number of tokens overall is similar to the document-aligned sample of the low-resource Wikipedia. This results in topic-adjusted Wikipedias consisting of 14.3M tokens for \textsc{es} and \textsc{eu}, 26.1M tokens for \textsc{es} and \textsc{gl}, and 409k tokens for \textsc{es} and \textsc{qu}. We additionally control for morphology by lemmatising the Wikipedia samples. For Spanish paired with each other language, we use training dictionaries that are similar in size and distribution. We learn monolingual embeddings on each subsampled Wikipedia corpus and align the resulting embeddings with English. We follow the same process to sample the Bengali Wikipedia to make its topical distribution aligned with the samples of the Urdu and Tamil Wikipedias: this results in topic-adjusted Wikipedias consisting of 3.8M tokens for Bengali--Urdu, and 8.1M word tokens for Bengali--Tamil.

The results are provided in Table~\ref{tab:topic-lemmas}. We observe that inequality in training resources accounts for a large part of the performance gap. Controlling for topical skew and morphology reduces the gap further and results in nearly identical performance for Spanish compared to Quechua and Galician, respectively. For Galician, lemmatisation slightly widens the gap, likely due to a weaker lemmatiser. For Basque, the remaining gap may be explained by the remaining typological differences between the two languages.\footnote{The drop in performance for \textsc{en--es} in the setup with full Wikipedias when we analyze Spanish and Quechua is attributed to a smaller and lower-quality training lexicon (to make it comparable to the corresponding \textsc{en}--\textsc{qu} lexicon).} We also observe similar patterns in experiments with \bn, \ur, and \ta in Table~\ref{tab:topic-lemmas}: training with comparable samples with additional morphological processing reduces the observed gap in performance between \textsc{en--bn} and \textsc{en--ta}, as well as between \textsc{en--bn} and \textsc{en--ur}. This again hints that other factors besides inherent language dissimilarity are at play and contribute to reduced isomorphism between embedding spaces.



\label{s:discussion}

\vspace{1.8mm}
\noindent \textbf{Does isomorphism increase beyond convergence?} 
In our experiments, we have measured how training monolingual word embeddings improves their isomorphism with embeddings in other languages. In doing so, we observed that \emph{ isomorphism increases even as validation (BLI) scores and training losses plateau } (see Figure ~\ref{fig:IB}). One possible explanation is that the random oscillations of SGD may lead the weights towards a high-entropy solution, which is more likely to be isomorphic. We highlight and discuss connections to work in optimization and generalization in the appendix. There, we also discuss more speculative implications and connections between (non-)isomorphism and vocabulary alignment across different languages.

\section{Conclusion}
\label{s:conclusion}
We have provided a series of analyses that demonstrate together that non-isomorphism is not---as previously assumed---primarily a result of typological differences between languages, but in large part due to degenerate vector spaces and discrepancies between monolingual training regimes and data availability. Through controlled experiments in simulated low-resource scenarios, also involving languages with different morphology that are spoken in culturally related regions, we found that such vector spaces mainly arise from poor conditioning during training. The study suggests that besides improving our alignment algorithms for distant languages \cite{Vulic2019}, we should also focus on improving monolingual word vector spaces, and monolingual training conditions to unlock a true potential of cross-lingual learning. 

\section*{Acknowledgments}
The work of IV is supported by the ERC Consolidator Grant LEXICAL: Lexical Acquisition Across Languages (no 648909). AS is supported by a Google Focused Research Award. We thank Chris Dyer and Phil Blunsom for feedback on a draft.

\bibliographystyle{acl_natbib}
\bibliography{ref}

\begin{thebibliography}{48}
\expandafter\ifx\csname natexlab\endcsname\relax\def\natexlab#1{#1}\fi

\bibitem[{Al-Rfou et~al.(2013)Al-Rfou, Perozzi, and
  Skiena}]{polyglot:2013:ACL-CoNLL}
Rami Al-Rfou, Bryan Perozzi, and Steven Skiena. 2013.
\newblock \href {http://www.aclweb.org/anthology/W13-3520} {Polyglot:
  Distributed word representations for multilingual nlp}.
\newblock In \emph{Proceedings of CoNLL 2013}, pages 183--192.

\bibitem[{Alvarez-Melis and Jaakkola(2018)}]{Alvarez-Melis2018gromov}
David Alvarez-Melis and Tommi~S. Jaakkola. 2018.
\newblock \href {https://www.aclweb.org/anthology/D18-1214.pdf}
  {{Gromov-Wasserstein} alignment of word embedding spaces}.
\newblock In \emph{Proceedings of EMNLP 2018}, pages 1881--1890.

\bibitem[{Ammar et~al.(2016)Ammar, Mulcaire, Tsvetkov, Lample, Dyer, and
  Smith}]{Ammar2016a}
Waleed Ammar, George Mulcaire, Yulia Tsvetkov, Guillaume Lample, Chris Dyer,
  and Noah~A. Smith. 2016.
\newblock \href {https://arxiv.org/abs/1602.01925} {Massively multilingual word
  embeddings}.
\newblock \emph{arXiv preprint arXiv:1602.01925}.

\bibitem[{Artetxe et~al.(2018{\natexlab{a}})Artetxe, Labaka, and
  Agirre}]{Artetxe2018}
Mikel Artetxe, Gorka Labaka, and Eneko Agirre. 2018{\natexlab{a}}.
\newblock \href {https://www.aclweb.org/anthology/P18-1073.pdf} {{A robust
  self-learning method for fully unsupervised cross-lingual mappings of word
  embeddings}}.
\newblock In \emph{Proceedings of ACL 2018}, pages 789--798.

\bibitem[{Artetxe et~al.(2018{\natexlab{b}})Artetxe, Labaka, and
  Agirre}]{Artetxe2018aaai}
Mikel Artetxe, Gorka Labaka, and Eneko Agirre. 2018{\natexlab{b}}.
\newblock \href
  {https://www.aaai.org/ocs/index.php/AAAI/AAAI18/paper/view/16935}
  {Generalizing and improving bilingual word embedding mappings with a
  multi-step framework of linear transformations}.
\newblock In \emph{Proceedings of AAAI 2018}, pages 5012--5019.

\bibitem[{Baroni et~al.(2014)Baroni, Dinu, and Kruszewski}]{Baroni2014}
Marco Baroni, Georgiana Dinu, and German Kruszewski. 2014.
\newblock \href {https://www.aclweb.org/anthology/P14-1023} {{Don't count,
  predict! A systematic comparison of context-counting vs. context-predicting
  semantic vectors}}.
\newblock In \emph{Proceedings of ACL 2014}, pages 238--247.

\bibitem[{Bojanowski et~al.(2017)Bojanowski, Grave, Joulin, and
  Mikolov}]{Bojanowski:2017tacl}
Piotr Bojanowski, Edouard Grave, Armand Joulin, and Tomas Mikolov. 2017.
\newblock \href {http://arxiv.org/abs/1607.04606} {Enriching word vectors with
  subword information}.
\newblock \emph{Transactions of the ACL}, 5:135--146.

\bibitem[{Chazal et~al.(2009)Chazal, Cohen-Steiner, Guibas, M{\'e}moli, and
  Oudot}]{chazal2009gromov}
Fr{\'e}d{\'e}ric Chazal, David Cohen-Steiner, Leonidas~J Guibas, Facundo
  M{\'e}moli, and Steve~Y Oudot. 2009.
\newblock \href
  {https://geometry.stanford.edu/papers/ccsgmo-ghsssp-09/ccsgmo-ghsssp-09.pdf}
  {Gromov-{H}ausdorff stable signatures for shapes using persistence}.
\newblock In \emph{Computer Graphics Forum}, volume~28, pages 1393--1403.

\bibitem[{Conneau et~al.(2020)Conneau, Khandelwal, Goyal, Chaudhary, Wenzek,
  Guzmán, Grave, Ott, Zettlemoyer, and Stoyanov}]{Conneau:2019arxiv}
Alexis Conneau, Kartikay Khandelwal, Naman Goyal, Vishrav Chaudhary, Guillaume
  Wenzek, Francisco Guzmán, Edouard Grave, Myle Ott, Luke Zettlemoyer, and
  Veselin Stoyanov. 2020.
\newblock \href {https://arxiv.org/pdf/1911.02116.pdf} {Unsupervised
  cross-lingual representation learning at scale}.
\newblock In \emph{Proceedings of ACL 2020}, pages 8440--8451.

\bibitem[{Conneau et~al.(2018)Conneau, Lample, Ranzato, Denoyer, and
  J{\'{e}}gou}]{Conneau2018}
Alexis Conneau, Guillaume Lample, Marc'Aurelio Ranzato, Ludovic Denoyer, and
  Herv{\'{e}} J{\'{e}}gou. 2018.
\newblock \href
  {https://doi.org/http://dx.doi.org/10.1111/j.1540-4560.2007.00543.x} {Word
  translation without parallel data}.
\newblock In \emph{Proceedings of ICLR 2018}.

\bibitem[{Czarnowska et~al.(2019)Czarnowska, Ruder, Grave, Cotterell, and
  Copestake}]{Czarnowska2019}
Paula Czarnowska, Sebastian Ruder, Edouard Grave, Ryan Cotterell, and Ann
  Copestake. 2019.
\newblock \href {https://www.aclweb.org/anthology/D19-1090} {Don't forget the
  long tail! {A} comprehensive analysis of morphological generalization in
  bilingual lexicon induction}.
\newblock In \emph{Proceedings of EMNLP 2019}, pages 974--983.

\bibitem[{Dubossarsky et~al.(2020)Dubossarsky, Vuli\'{c}, Reichart, and
  Korhonen}]{Dubossarsky:2020emnlp}
Haim Dubossarsky, Ivan Vuli\'{c}, Roi Reichart, and Anna Korhonen. 2020.
\newblock The secret is in the spectra: {P}redicting cross-lingual task
  performance with spectral similarity measures.
\newblock In \emph{Proceedings of EMNLP 2020}.

\bibitem[{Fujinuma et~al.(2019)Fujinuma, Boyd-Graber, and Paul}]{Fujinuma2019}
Yoshinari Fujinuma, Jordan Boyd-Graber, and Michael~J. Paul. 2019.
\newblock \href {https://www.aclweb.org/anthology/P19-1489} {A resource-free
  evaluation metric for cross-lingual word embeddings based on graph
  modularity}.
\newblock In \emph{Proceedings of ACL 2019}, pages 4952--4962.

\bibitem[{Gerz et~al.(2016)Gerz, Vuli{\'c}, Hill, Reichart, and
  Korhonen}]{Gerz:2016emnlp}
Daniela Gerz, Ivan Vuli{\'c}, Felix Hill, Roi Reichart, and Anna Korhonen.
  2016.
\newblock \href {https://www.aclweb.org/anthology/D16-1235} {{S}im{V}erb-3500:
  {A} large-scale evaluation set of verb similarity}.
\newblock In \emph{Proceedings of EMNLP 2016}, pages 2173--2182.

\bibitem[{Glavaš et~al.(2019)Glavaš, Litschko, Ruder, and
  Vuli{\'{c}}}]{Glavas2019}
Goran Glavaš, Robert Litschko, Sebastian Ruder, and Ivan Vuli{\'{c}}. 2019.
\newblock \href {https://doi.org/10.18653/v1/P19-1070} {How to (properly)
  evaluate cross-lingual word embeddings: On strong baselines, comparative
  analyses, and some misconceptions}.
\newblock In \emph{Proceedings of ACL 2019}, pages 710--721.

\bibitem[{Grave et~al.(2018)Grave, Bojanowski, Gupta, Joulin, and
  Mikolov}]{Grave:2018lrec}
Edouard Grave, Piotr Bojanowski, Prakhar Gupta, Armand Joulin, and Tomas
  Mikolov. 2018.
\newblock \href {http://aclweb.org/anthology/L18-1550} {Learning word vectors
  for 157 languages}.
\newblock In \emph{Proceedings of LREC 2018}, pages 3483--3487.

\bibitem[{Hartmann et~al.(2018)Hartmann, Kementchedjhieva, and
  S{\o}gaard}]{Hartmann2018}
Mareike Hartmann, Yova Kementchedjhieva, and Anders S{\o}gaard. 2018.
\newblock \href {https://doi.org/10.18653/v1/D18-1056} {{Why is unsupervised
  alignment of English embeddings from different algorithms so hard?}}
\newblock In \emph{Proceedings of EMNLP 2018}, pages 582--586.

\bibitem[{Hartmann et~al.(2019)Hartmann, Kementchedjhieva, and
  S{\o}gaard}]{Hartmann:2019neurips}
Mareike Hartmann, Yova Kementchedjhieva, and Anders S{\o}gaard. 2019.
\newblock \href
  {http://papers.nips.cc/paper/8836-comparing-unsupervised-word-translation-methods-step-by-step}
  {Comparing unsupervised word translation methods step by step}.
\newblock In \emph{Proceedings of NeurIPS 2019}, pages 6031--6041.

\bibitem[{Hill et~al.(2015)Hill, Reichart, and Korhonen}]{hill2015simlex}
Felix Hill, Roi Reichart, and Anna Korhonen. 2015.
\newblock \href {https://www.aclweb.org/anthology/J15-4004.pdf} {Simlex-999:
  Evaluating semantic models with (genuine) similarity estimation}.
\newblock \emph{Computational Linguistics}, 41(4):665--695.

\bibitem[{Kamholz et~al.(2014)Kamholz, Pool, and Colowick}]{Kamholz2014panlex}
David Kamholz, Jonathan Pool, and Susan~M. Colowick. 2014.
\newblock \href
  {http://www.lrec-conf.org/proceedings/lrec2014/summaries/1029.html}
  {Pan{L}ex: {B}uilding a resource for panlingual lexical translation}.
\newblock In \emph{Proceedings of LREC 2014}, pages 3145--3150.

\bibitem[{Kementchedjhieva et~al.(2019)Kementchedjhieva, Hartmann, and
  S{\o}gaard}]{Kementchedjhieva2019}
Yova Kementchedjhieva, Mareike Hartmann, and Anders S{\o}gaard. 2019.
\newblock \href {https://doi.org/10.18653/v1/D19-1328} {Lost in evaluation:
  {M}isleading benchmarks for bilingual dictionary induction}.
\newblock In \emph{Proceedings of EMNLP 2019}, pages 3336--3341.

\bibitem[{Majid(2010)}]{Majid:10}
Asifa Majid. 2010.
\newblock \href
  {https://www.oxfordhandbooks.com/view/10.1093/oxfordhb/9780199641604.001.0001/oxfordhb-9780199641604-e-020}
  {Comparing lexicons cross-linguistically}.
\newblock In \emph{Oxford Handbook of the Word}.

\bibitem[{Mandera et~al.(2017)Mandera, Keuleers, and Brysbaert}]{Mandera:2017}
Pawe{\l} Mandera, Emmanuel Keuleers, and Marc Brysbaert. 2017.
\newblock \href {https://biblio.ugent.be/publication/8502464} {Explaining human
  performance in psycholinguistic tasks with models of semantic similarity
  based on prediction and counting: {A} review and empirical validation}.
\newblock \emph{Journal of Memory and Language}, 92:57--78.

\bibitem[{Mikolov et~al.(2013{\natexlab{a}})Mikolov, Le, and
  Sutskever}]{Mikolov2013exploiting}
Tomas Mikolov, Quoc~V. Le, and Ilya Sutskever. 2013{\natexlab{a}}.
\newblock \href {https://arxiv.org/abs/1309.4168} {Exploiting similarities
  among languages for machine translation}.
\newblock \emph{arXiv preprint arXiv:1309.4168}.

\bibitem[{Mikolov et~al.(2013{\natexlab{b}})Mikolov, Sutskever, Chen, Corrado,
  and Dean}]{Mikolov:2013nips}
Tomas Mikolov, Ilya Sutskever, Kai Chen, Gregory~S. Corrado, and Jeffrey Dean.
  2013{\natexlab{b}}.
\newblock \href
  {https://papers.nips.cc/paper/5021-distributed-representations-of-words-and-phrases-and-their-compositionality.pdf}
  {Distributed representations of words and phrases and their
  compositionality}.
\newblock In \emph{Proceedings of NeurIPS 2013}, pages 3111--3119.

\bibitem[{Mikolov et~al.(2013{\natexlab{c}})Mikolov, Yih, and
  Zweig}]{Mikolov2013naacl}
Tomas Mikolov, Wen-tau Yih, and Geoffrey Zweig. 2013{\natexlab{c}}.
\newblock \href {https://www.aclweb.org/anthology/N13-1090.pdf} {Linguistic
  regularities in continuous space word representations}.
\newblock In \emph{Proceedings of NAACL-HLT 2013}, pages 746--751.

\bibitem[{Nakashole(2018)}]{Nakashole2018}
Ndapa Nakashole. 2018.
\newblock \href {https://doi.org/10.18653/v1/D18-1047} {{NORMA: N}eighborhood
  sensitive maps for multilingual word embeddings}.
\newblock In \emph{Proceedings of EMNLP 2018}, pages 512--522.

\bibitem[{Ormazabal et~al.(2019)Ormazabal, Artetxe, Labaka, Soroa, and
  Agirre}]{Ormazabal2019}
Aitor Ormazabal, Mikel Artetxe, Gorka Labaka, Aitor Soroa, and Eneko Agirre.
  2019.
\newblock \href {https://doi.org/10.18653/v1/P19-1492} {Analyzing the
  limitations of cross-lingual word embedding mappings}.
\newblock In \emph{Proceedings of ACL 2019}, pages 4990--4995.

\bibitem[{Patra et~al.(2019)Patra, Moniz, Garg, Gormley, and
  Neubig}]{Patra2019}
Barun Patra, Joel Ruben~Antony Moniz, Sarthak Garg, Matthew~R. Gormley, and
  Graham Neubig. 2019.
\newblock \href {https://doi.org/10.18653/v1/P19-1018} {Bilingual lexicon
  induction with semi-supervision in non-isometric embedding spaces}.
\newblock In \emph{Proceedings of ACL 2019}, pages 184--193.

\bibitem[{Pavlenko(2009)}]{Pavlenko:09}
Aneta Pavlenko, editor. 2009.
\newblock \href
  {http://www.anetapavlenko.com/pdf/Chapter_6_Conceptual_Representation_in_the_Bilingual_Lexicon_and_Second_Language_Vocabulary_Learning.pdf}
  {\emph{The Bilingual Mental Lexicon}}.

\bibitem[{Pennington et~al.(2014)Pennington, Socher, and
  Manning}]{Pennington:2014emnlp}
Jeffrey Pennington, Richard Socher, and Christopher~D. Manning. 2014.
\newblock \href {https://doi.org/10.3115/v1/D14-1162} {{Glove: G}lobal vectors
  for word representation}.
\newblock In \emph{Proceedings of EMNLP 2014}, pages 1532--1543.

\bibitem[{Ravfogel et~al.(2019)Ravfogel, Goldberg, and Linzen}]{Ravfogel2019}
Shauli Ravfogel, Yoav Goldberg, and Tal Linzen. 2019.
\newblock \href {https://doi.org/10.18653/v1/N19-1356} {Studying the inductive
  biases of {RNNs} with synthetic variations of natural languages}.
\newblock In \emph{Proceedings of NAACL-HLT 2019}, pages 3532--3542.

\bibitem[{Ruder et~al.(2019)Ruder, Vuli{\'{c}}, and
  S{\o}gaard}]{Ruder2019survey}
Sebastian Ruder, Ivan Vuli{\'{c}}, and Anders S{\o}gaard. 2019.
\newblock \href {https://arxiv.org/pdf/1706.04902.pdf} {A survey of
  cross-lingual word embedding models}.
\newblock \emph{Journal of Artificial Intelligence Research}, 65:569--631.

\bibitem[{Sahlgren and Lenci(2016)}]{Sahlgren2016}
Magnus Sahlgren and Alessandro Lenci. 2016.
\newblock \href {https://www.aclweb.org/anthology/D16-1099} {{The effects of
  data size and frequency range on distributional semantic models}}.
\newblock In \emph{Proceedings of EMNLP 2016}, pages 975--980.

\bibitem[{Shwartz-Ziv and Tishby(2017)}]{shwartz2017opening}
Ravid Shwartz-Ziv and Naftali Tishby. 2017.
\newblock \href {https://arxiv.org/abs/1703.00810} {Opening the black box of
  deep neural networks via information}.
\newblock \emph{arXiv preprint arXiv:1703.00810}.

\bibitem[{S{\o}gaard et~al.(2018)S{\o}gaard, Ruder, and
  Vuli{\'c}}]{sogaard2018limitations}
Anders S{\o}gaard, Sebastian Ruder, and Ivan Vuli{\'c}. 2018.
\newblock \href {https://doi.org/10.18653/v1/P18-1072} {On the limitations of
  unsupervised bilingual dictionary induction}.
\newblock In \emph{Proceedings of ACL 2018}, pages 778--788.

\bibitem[{Straka and Strakov{\'a}(2017)}]{straka2017tokenizing}
Milan Straka and Jana Strakov{\'a}. 2017.
\newblock \href {https://www.aclweb.org/anthology/K17-3009.pdf} {Tokenizing,
  {POS} tagging, lemmatizing and parsing {UD2.0 with UDPipe}}.
\newblock In \emph{Proceedings of the CoNLL 2017 Shared Task: Multilingual
  Parsing from Raw Text to Universal Dependencies}, pages 88--99.

\bibitem[{Thompson et~al.(2018)Thompson, Roberts, and Lupyan}]{Thompson:ea:18}
Bill Thompson, Sean Roberts, and Gary Lupyan. 2018.
\newblock \href
  {https://pure.mpg.de/rest/items/item_2622749_4/component/file_2622747/content}
  {Quantifying semantic alignment across languages}.
\newblock \emph{MPI Tech Report}.

\bibitem[{Torabi~Asr et~al.(2018)Torabi~Asr, Zinkov, and Jones}]{Asr:2018naacl}
Fatemeh Torabi~Asr, Robert Zinkov, and Michael Jones. 2018.
\newblock \href {https://www.aclweb.org/anthology/N18-1062} {Querying word
  embeddings for similarity and relatedness}.
\newblock In \emph{Proceedings of NAACL-HLT 2018}, pages 675--684.

\bibitem[{Tsvetkov et~al.(2015)Tsvetkov, Faruqui, Ling, Lample, and
  Dyer}]{Tsvetkov2015}
Yulia Tsvetkov, Manaal Faruqui, Wang Ling, Guillaume Lample, and Chris Dyer.
  2015.
\newblock \href {https://doi.org/10.18653/v1/D15-1243} {Evaluation of word
  vector representations by subspace alignment}.
\newblock In \emph{Proceedings of EMNLP 2015}, pages 2049--2054.

\bibitem[{Vossen(2002)}]{Vossen:02}
Piet Vossen. 2002.
\newblock \href
  {https://www.cairn.info/revue-francaise-de-linguistique-appliquee-2002-1-page-27.htm}
  {{WordNet, EuroWordNet and Global WordNet }}.
\newblock \emph{Revue fran\c{c}aise de linguistique appliqu\'{e}e}, VII.

\bibitem[{Vuli{\'{c}} et~al.(2019)Vuli{\'{c}}, Glava{\v{s}}, Reichart, and
  Korhonen}]{Vulic2019}
Ivan Vuli{\'{c}}, Goran Glava{\v{s}}, Roi Reichart, and Anna Korhonen. 2019.
\newblock \href {https://doi.org/10.18653/v1/D19-1449} {Do we really need fully
  unsupervised cross-lingual embeddings?}
\newblock In \emph{Proceedings of EMNLP 2019}, pages 4407--4418.

\bibitem[{Xing et~al.(2015)Xing, Liu, Wang, and Lin}]{Xing2015orthogonal}
Chao Xing, Chao Liu, Dong Wang, and Yiye Lin. 2015.
\newblock \href {https://doi.org/10.3115/v1/N15-1104} {Normalized word
  embedding and orthogonal transform for bilingual word translation}.
\newblock In \emph{Proceedings of NAACL-HLT 2015}, pages 1005--1010.

\bibitem[{Xu et~al.(2018)Xu, Yang, Otani, and Wu}]{Xu2018}
Ruochen Xu, Yiming Yang, Naoki Otani, and Yuexin Wu. 2018.
\newblock \href {https://doi.org/10.18653/v1/D18-1268} {Unsupervised
  cross-lingual transfer of word embedding spaces}.
\newblock In \emph{Proceedings of EMNLP 2018}, pages 2465--2474.

\bibitem[{Yaghoobzadeh and Sch{\"{u}}tze(2016)}]{Yaghoobzadeh2016}
Yadollah Yaghoobzadeh and Hinrich Sch{\"{u}}tze. 2016.
\newblock \href {https://doi.org/10.18653/v1/P16-1023} {Intrinsic subspace
  evaluation of word embedding representations}.
\newblock In \emph{Proceedings of ACL 2016}, pages 236--246.

\bibitem[{Zhang et~al.(2017)Zhang, Liu, Luan, and Sun}]{Zhang2017earthmover}
Meng Zhang, Yang Liu, Huanbo Luan, and Maosong Sun. 2017.
\newblock \href {https://doi.org/10.18653/v1/D17-1207} {Earth mover's distance
  minimization for unsupervised bilingual lexicon induction}.
\newblock In \emph{Proceedings of EMNLP 2017}, pages 1934--1945.

\bibitem[{Zhang et~al.(2019)Zhang, Xu, Kawarabayashi, Jegelka, and
  Boyd-Graber}]{Zhang2019}
Mozhi Zhang, Keyulu Xu, Ken-ichi Kawarabayashi, Stefanie Jegelka, and Jordan
  Boyd-Graber. 2019.
\newblock \href {https://doi.org/10.18653/v1/P19-1307} {Are girls neko or
  shojo? {C}ross-lingual alignment of non-isomorphic embeddings with iterative
  normalization}.
\newblock In \emph{Proceedings of ACL 2019}, pages 3180--3189.

\bibitem[{Zhang et~al.(2018)Zhang, Saxe, Advani, and Lee}]{Zhang:ea:18}
Yao Zhang, Andrew Saxe, Madhu Advani, and Alpha Lee. 2018.
\newblock \href {https://arxiv.org/abs/1803.01927} {Energy–entropy
  competition and the effectiveness of stochastic gradient descent in machine
  learning}.
\newblock \emph{Molecular Physics}, 116.

\end{thebibliography}

\appendix
\clearpage
\label{sec:supplemental}
\section{Further Discussion}
\subsection{Does isomorphism increase beyond convergence?}

Recent studies of learning dynamics in deep neural nets observe that flatter optima generalise better than sharp optima \cite{Zhang:ea:18}: intuitively, it is because sharp minima correspond to more
complex, likely over-fitted, models. \newcite{Zhang:ea:18} show that analogous to the energy-entropy competition in statistical physics, wide but shallow minima can be optimal if the system is undersampled. 
SGD is assumed to generalise well because its inherent anisotropic noise biases it towards higher entropy minima. We hypothesise a similar explanation of our observations in terms of energy-entropy competition. Once loss is minimised, the random oscillations due to SGD noise lead the weights toward a high-entropy solution. We hypothesise monolingual high-entropy minima are more likely to be isomorphic. A related possible explanation is that the increased isomorphism results from model compression. This is analogous to the idea of two-phase learning \cite{shwartz2017opening}, whereby the initial fast convergence of SGD
is related to sufficiency of the representation, while the later asymptotic phase is related
to compression of the activations. 

\subsection{Do vocabularies align?} If languages reflect the world, they should convey semantic knowledge in a similar way, and it is therefore reasonable to assume that with enough data, induced word embeddings should be isomorphic. On the other hand, if languages impose structure on our conceptualisation of the world, non-isomorphic word embeddings could easily arise. Studies that engage with speakers of different languages in the real world \cite{Majid:10} are naturally limited in scope. 
Large-scale studies, on the other hand, have generally relied on distributional methods \cite{Thompson:ea:18}, leading to a chicken-and-egg scenario. \newcite{Vossen:02} discuss mismatches between WordNets across languages, including examples of hypernyms without translation equivalents, e.g., {\it dedo} in Spanish ({\em fingers} and {\it toes} in English). Such examples break isomorphism between languages, but are relatively rare. Another approach to the question of vocabulary alignment is to study lexical organisation in bilinguals and how it differs from that of monolingual speakers \cite{Pavlenko:09}.
While this paper obviously does not provide hard evidence for or against Sapir-Whorf-like hypotheses, our results suggest that the variation observed in BLI performance cannot trivially be attributed only to linguistic differences. 

\section{Additional Experiments}
\begin{figure*}[!t]
    \centering
    \begin{subfigure}[t]{0.328\linewidth}
        \centering
        \includegraphics[width=0.98\linewidth]{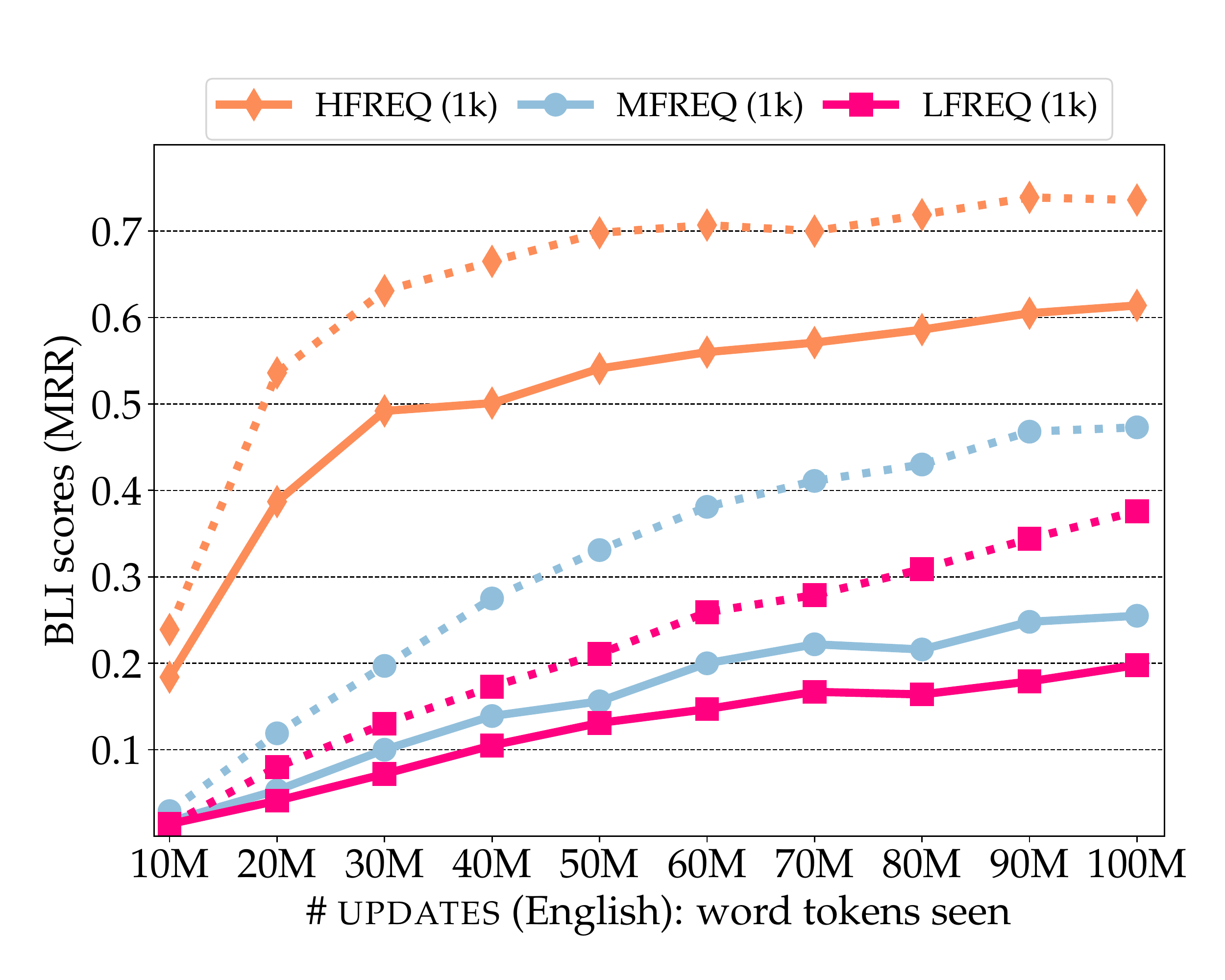}
        \caption{Early updates: \textsc{en} (1k)}
        \label{fig:en-early1k}
    \end{subfigure}
    \begin{subfigure}[t]{0.328\textwidth}
        \centering
        \includegraphics[width=0.98\linewidth]{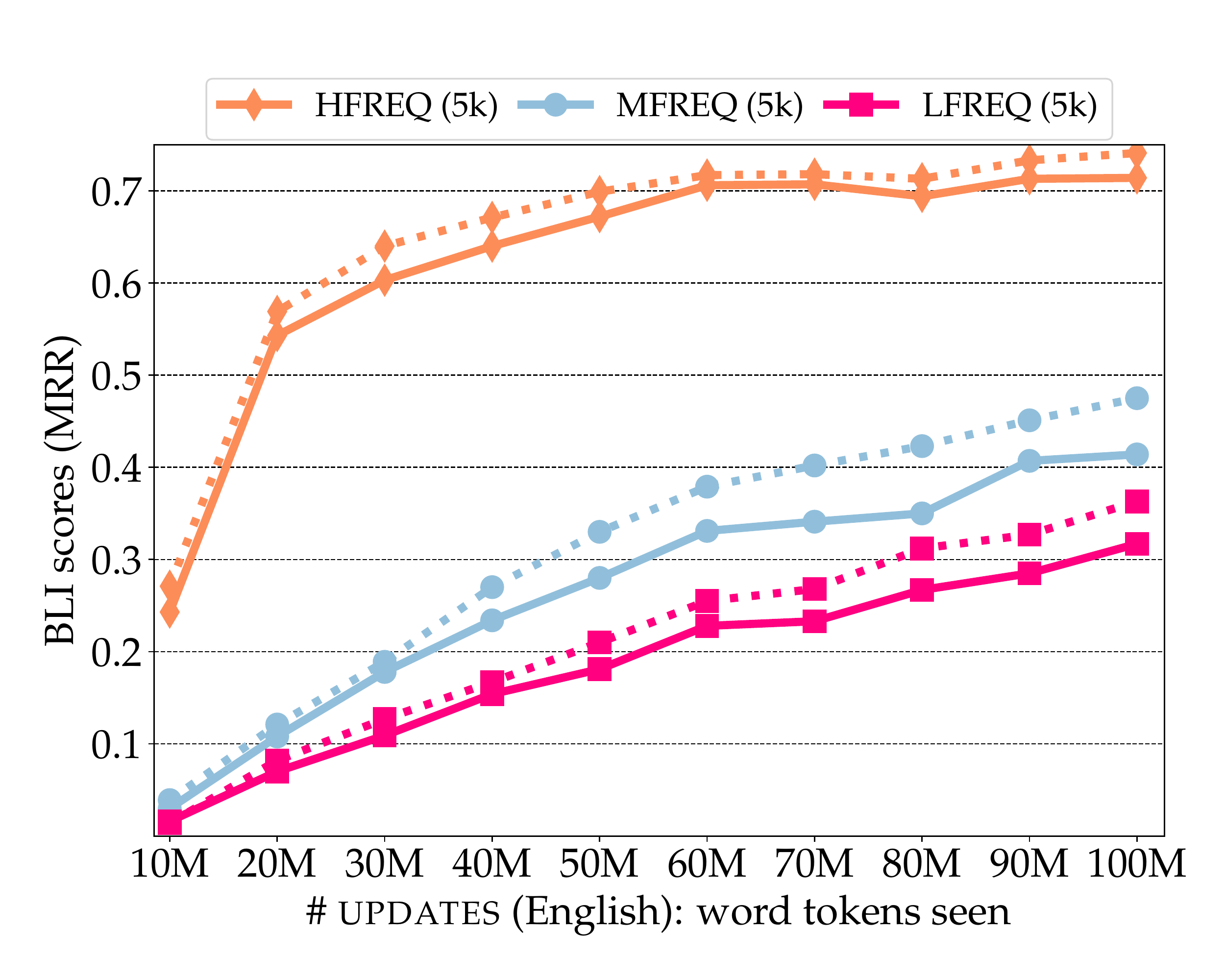}
        \caption{Early updates: \textsc{en} (5k)}
        \label{fig:en-early5k}
    \end{subfigure}
        \begin{subfigure}[t]{0.328\textwidth}
        \centering
        \includegraphics[width=0.98\linewidth]{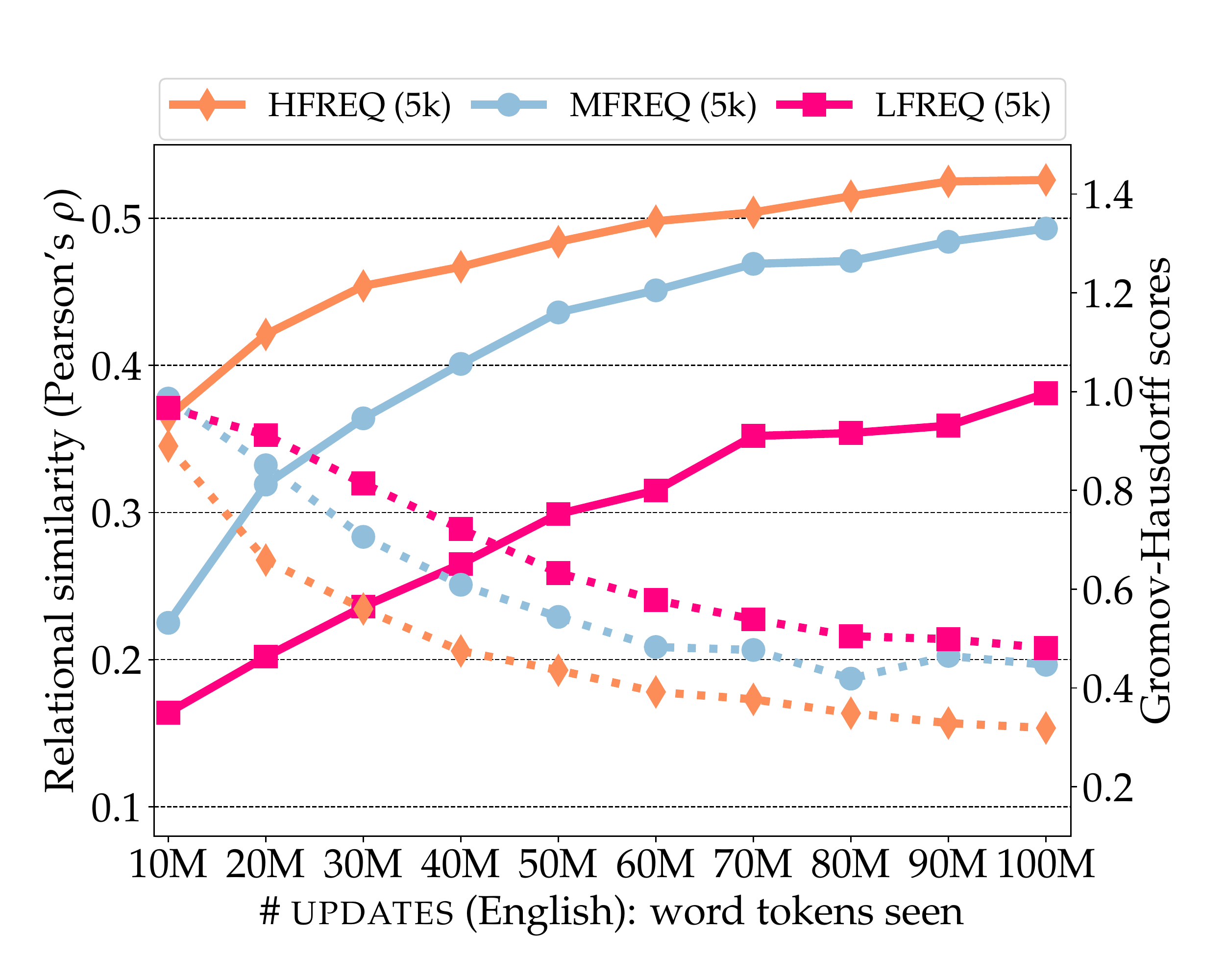}
        \caption{Early updates: \textsc{en} (isomorphism)}
        \label{fig:en-earlyiso}
    \end{subfigure}
    \vspace{-2mm}
    \caption{Impact of training duration on BLI and isomorphism, with a focus on the \textit{early training stages}. BLI scores (a+b) and isomorphism (c) measures of aligning a partially trained \textsc{en} vector space, where snapshots are taken after seeing $N$ word tokens in training, to a fully trained \textsc{es} vector space with a seed dictionary of 1k words (a) and 5k words (b) on the three evaluation sets representing different frequency bins. (c) shows how embedding spaces become more isomorphic over the course of training as measured by second-order similarity (on different frequency bins; solid lines, higher is better) and by Gromov-Hausdorff distance (dotted lines of the same colour and with the same symbols; lower is better).}
    \vspace{-1.5mm}
\label{fig:en-training-duration-early}
\end{figure*}

\begin{figure*}[!t]
    \centering
    \begin{subfigure}[t]{0.328\linewidth}
        \centering
        \includegraphics[width=0.98\linewidth]{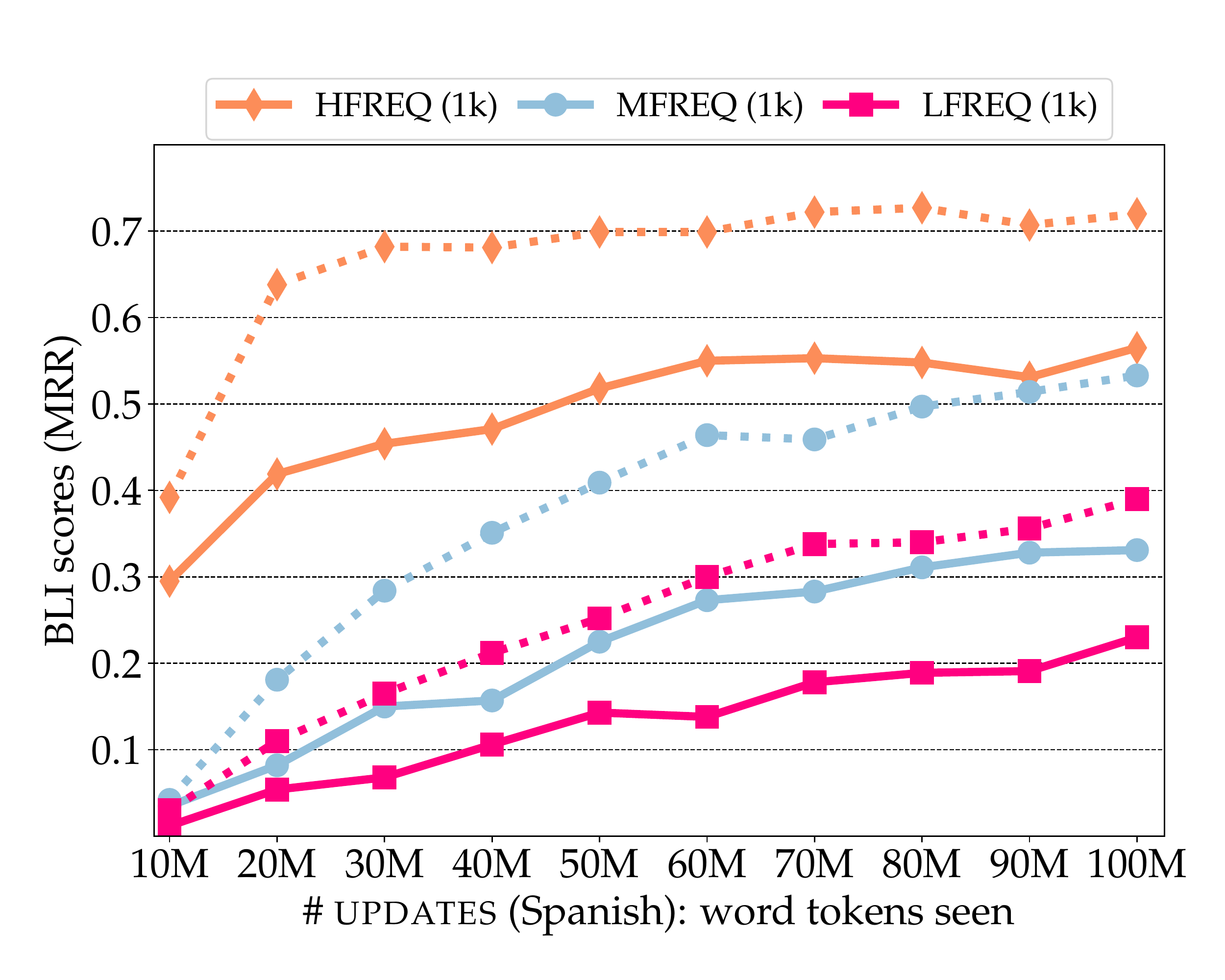}
        \caption{Early updates: \textsc{en} (1k)}
        \label{fig:en-early1k}
    \end{subfigure}
    \begin{subfigure}[t]{0.328\textwidth}
        \centering
        \includegraphics[width=0.98\linewidth]{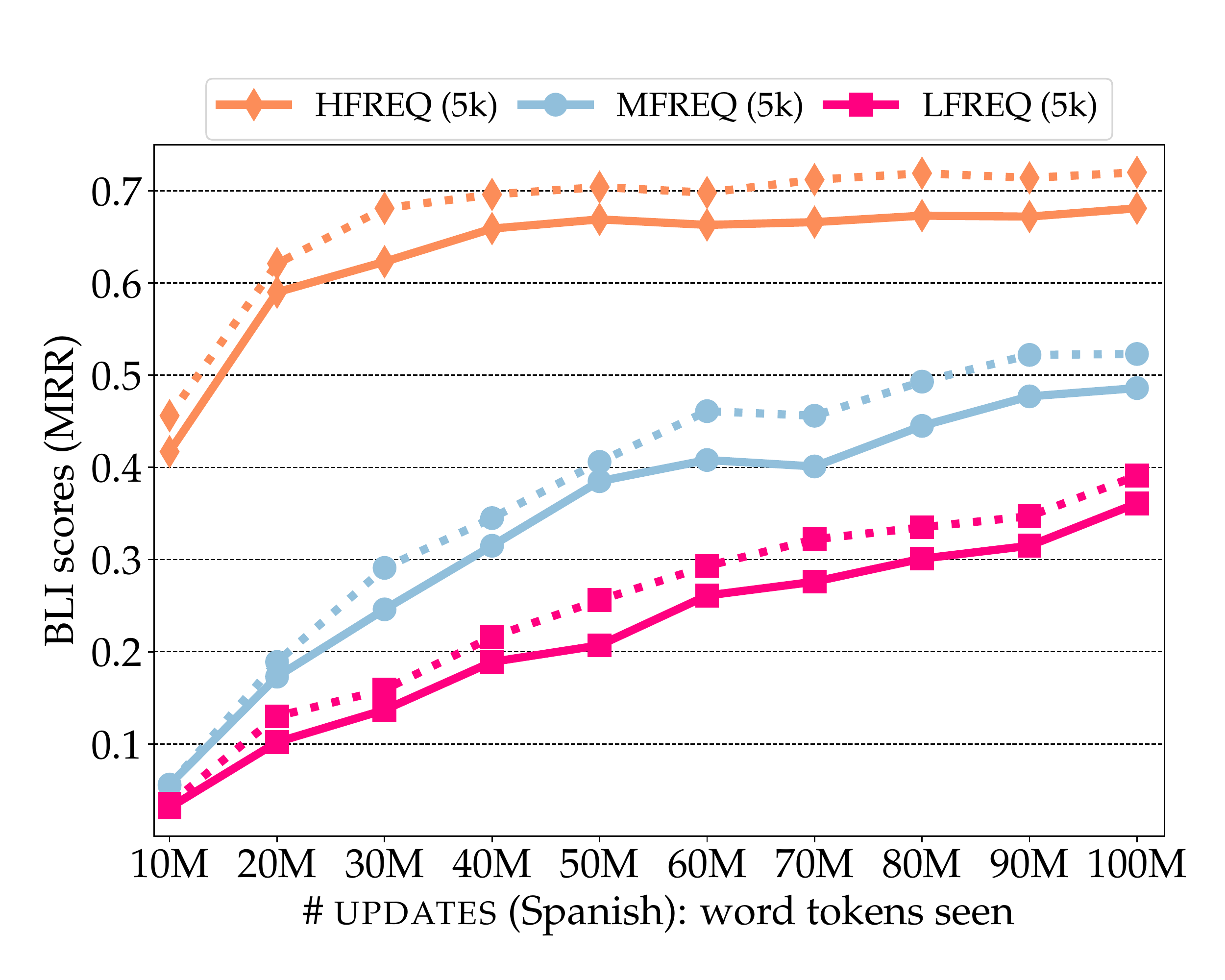}
        \caption{Early updates: \textsc{en} (5k)}
        \label{fig:en-early5k}
    \end{subfigure}
        \begin{subfigure}[t]{0.328\textwidth}
        \centering
        \includegraphics[width=0.98\linewidth]{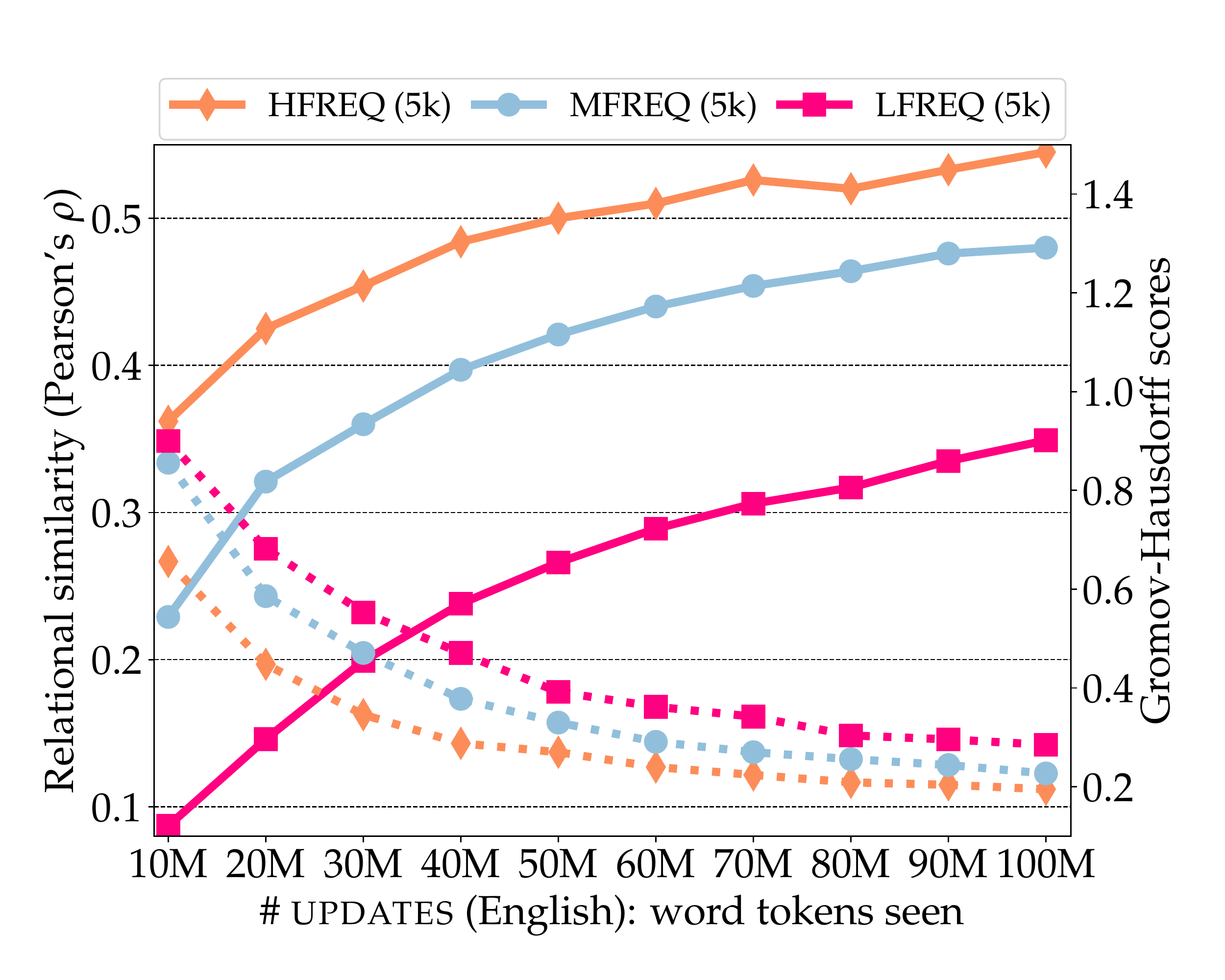}
        \caption{Early updates: \textsc{en} (isomorphism)}
        \label{fig:en-earlyiso}
    \end{subfigure}
    \vspace{-2mm}
    \caption{Impact of training duration on BLI and isomorphism, with a focus on the \textit{early training stages}. BLI scores (a+b) and isomorphism (c) measures of aligning a partially trained \textsc{es} vector space, where snapshots are taken after seeing $N$ word tokens in training, to a fully trained \textsc{en} vector space with a seed dictionary of 1k words (a) and 5k words (b) on the three evaluation sets representing different frequency bins. (c) shows how embedding spaces become more isomorphic over the course of training as measured by second-order similarity (on different frequency bins; solid lines, higher is better) and by Gromov-Hausdorff distance (dotted lines of the same colour and with the same symbols; lower is better).}
    \vspace{-1.5mm}
\label{fig:es-training-duration-early}
\end{figure*}

\begin{figure*}[t]
    \centering
    \begin{subfigure}[t]{0.408\linewidth}
        \centering
        \includegraphics[width=0.98\linewidth]{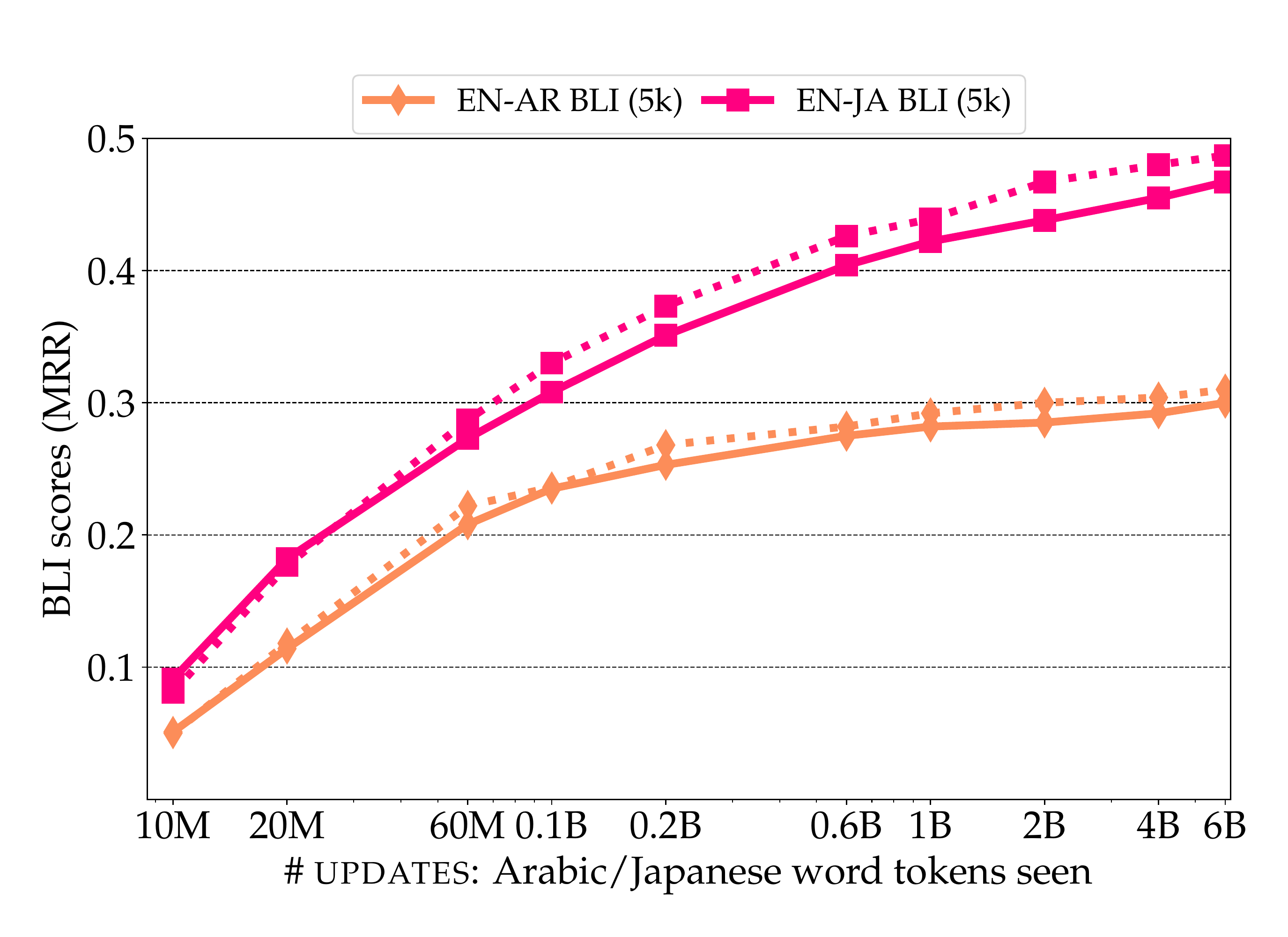}
        \caption{Updates: \textsc{ar} and \textsc{ja}}
        \label{fig:updates-arja-5k}
    \end{subfigure}
    \begin{subfigure}[t]{0.408\textwidth}
        \centering
        \includegraphics[width=0.98\linewidth]{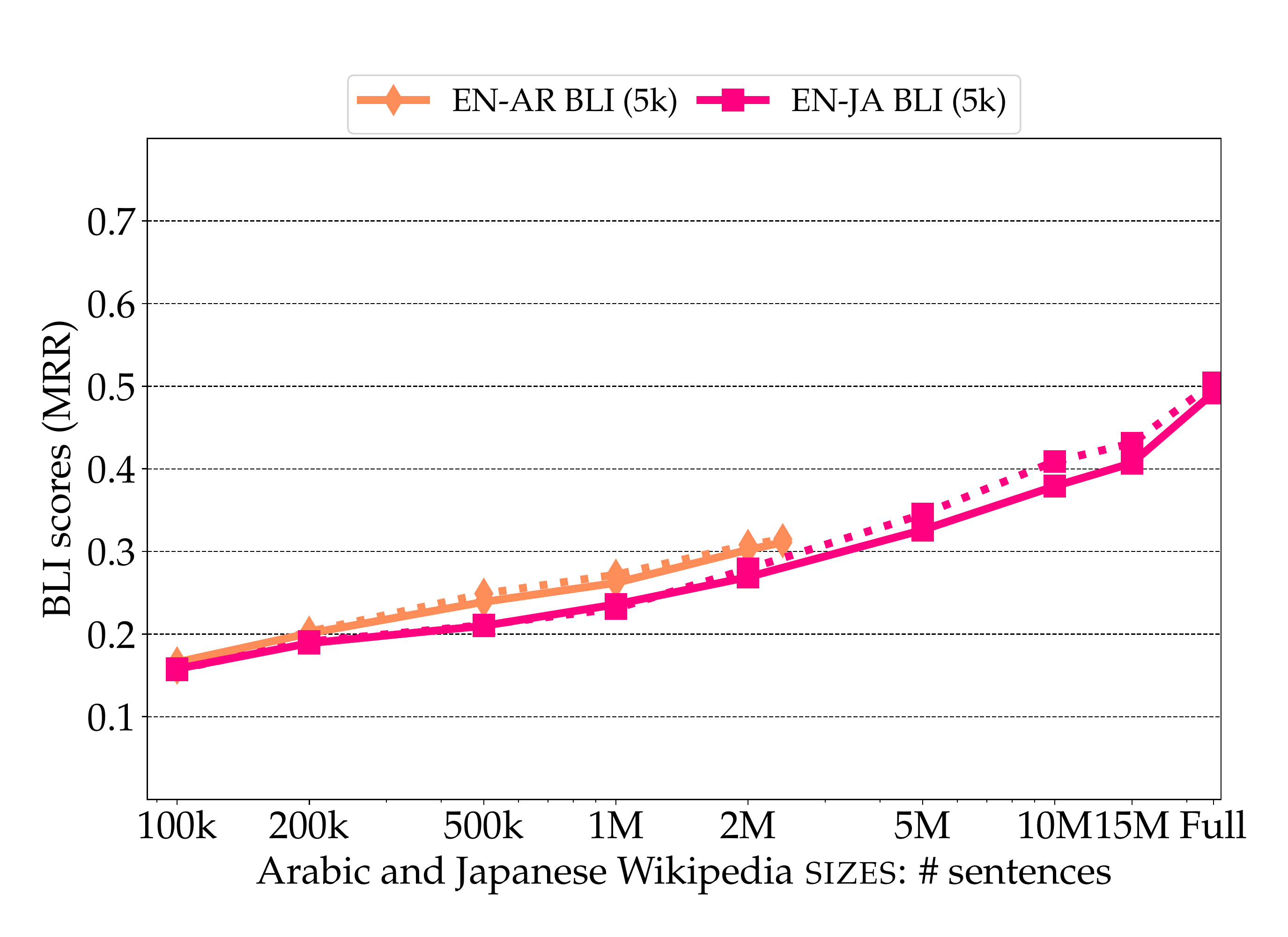}
        \caption{Different Wiki sizes: \textsc{ar} and \textsc{ja}}
        \label{fig:sizes-arja-5k}
    \end{subfigure}
    \vspace{-2mm}
    \caption{\textsc{en}--\textsc{ar/ja} BLI scores on the MUSE BLI benchmark relying on 5k seed pairs for learning the alignment. \textbf{(a)} Results with partially trained \textsc{ar} and \textsc{ja} vector spaces where snapshots are taken after $M$ updates (i.e., impact of training duration); \textbf{b)} Results with \textsc{ar} and \textsc{ja} vector spaces induced from data samples of different sizes (i.e., impact of dataset size). See the main paper for BLI scores with 5k seed translation pairs.}
    \vspace{-1.5mm}
\label{fig:arja-bli}
\end{figure*}

\begin{figure*}[t]
    \centering
    \begin{subfigure}[t]{0.408\linewidth}
        \centering
        \includegraphics[width=0.98\linewidth]{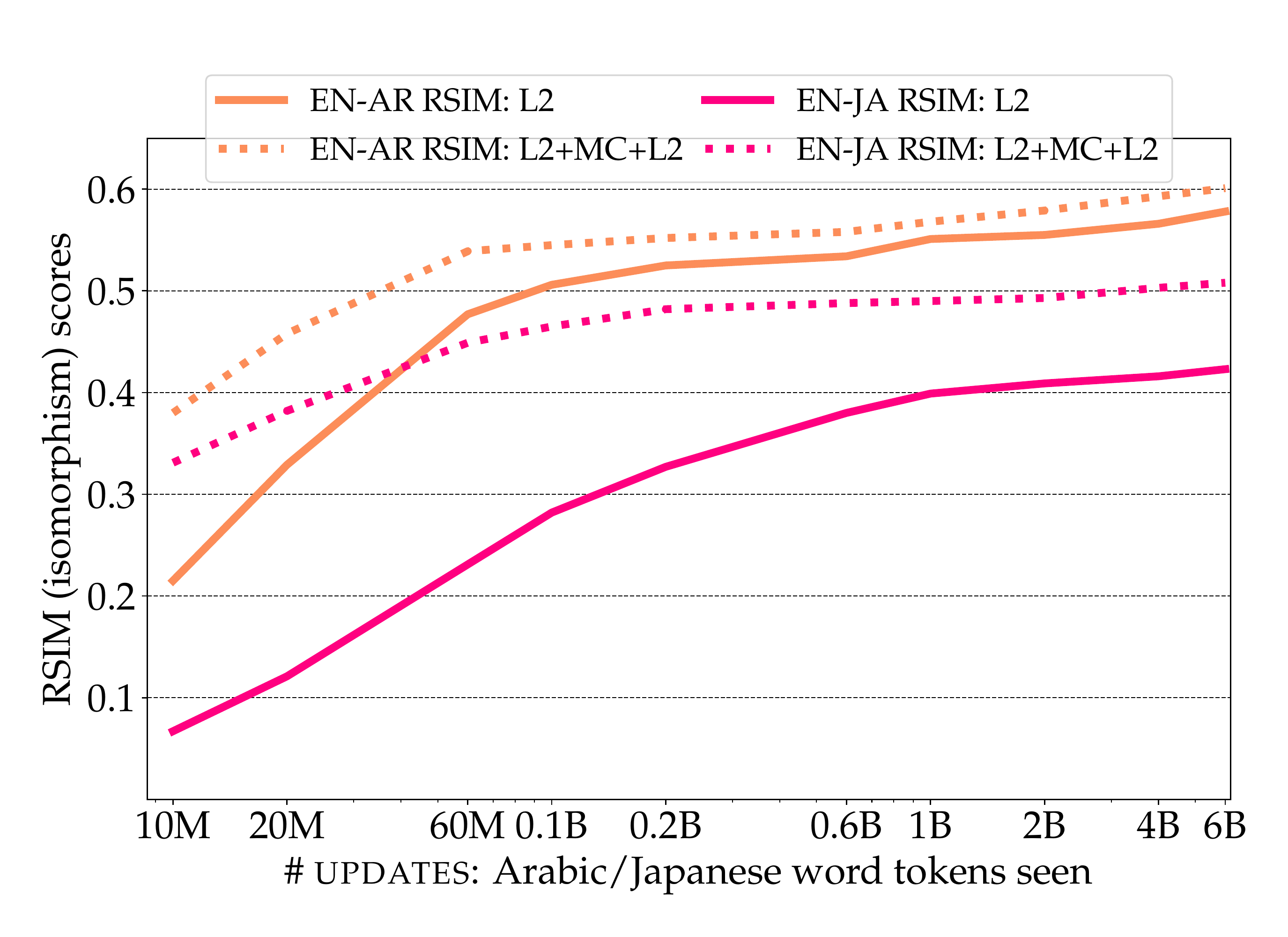}
        \caption{Updates: \textsc{ar} and \textsc{ja}}
        \label{fig:updates-arja-rsim}
    \end{subfigure}
    \begin{subfigure}[t]{0.408\textwidth}
        \centering
        \includegraphics[width=0.98\linewidth]{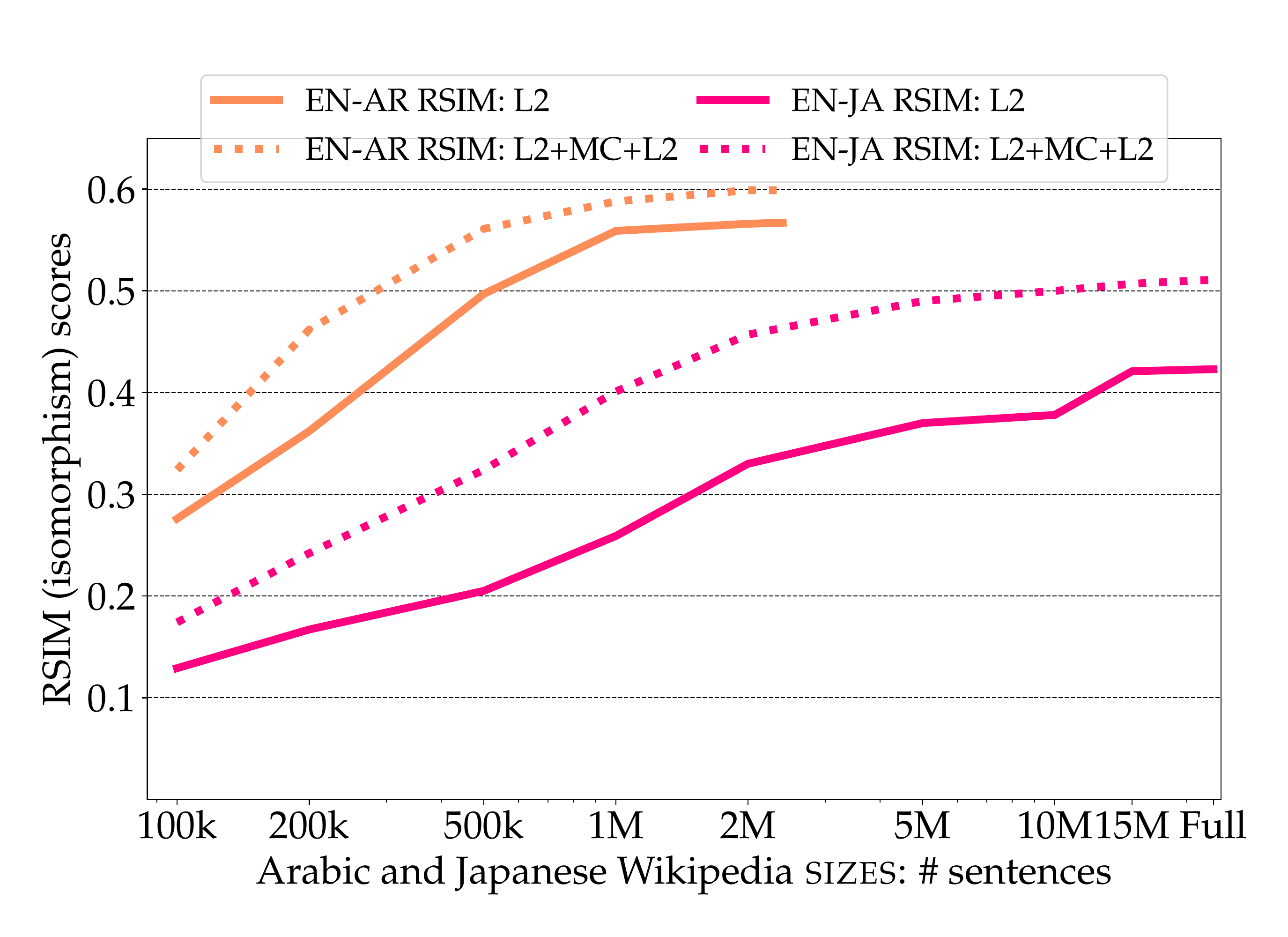}
        \caption{Different Wiki sizes: \textsc{ar} and \textsc{ja}}
        \label{fig:sizes-arja-rsim}
    \end{subfigure}
    \vspace{-2mm}
    \caption{The impact of \textbf{(a)} training duration and \textbf{(b)} dataset size on \textsc{en}--\textsc{ar/ja} isomorphism scores, also showing the impact of vector space preprocessing steps. We report the RSIM measure (higher is better, i.e., more isomorphic).}
    \vspace{-1.5mm}
\label{fig:arja-rsim}
\end{figure*}

Additional experiments that further support the main claims of the paper have been relegated to the appendix for clarity and compactness of presentation. We provide the following additional information:

\begin{itemize}
\item {\bf Table~\ref{tab:full-training}}. It provides ``reference'' BLI scores and scores stemming from isomorphism measures when we align fully trained EN and ES spaces, that is, when we rely on standard 15 epochs of fastText training on respective Wikipedias.

\item {\bf Figure~\ref{fig:en-training-duration-early}} and {\bf Figure~\ref{fig:es-training-duration-early}} show BLI and isomorphism scores at very early stages of training, both for EN and ES. In other words, one vector space is fully trained, while we take early-training snapshots (after seeing only 10M, 20M, \ldots, 100M word tokens in training) of the other vector space. The results again stress the importance of training corpus size as well as training duration---early training stages clearly lead to suboptimal performance and non-isomorphic spaces. However, such shorter training durations (in terms of the number of tokens) are often encountered ``in the wild'' with low-resource languages.

\item {\bf Figure~\ref{fig:updates-arja-5k}} and {\bf Figure~\ref{fig:sizes-arja-5k}} show the results with 5k seed translation pairs in different training regimes for \textsc{en-ar} and \textsc{en-ja} experiments. The results with 1k seed translation pairs are provided in the main paper.

\item {\bf Figure~\ref{fig:updates-arja-rsim}} and {\bf Figure~\ref{fig:sizes-arja-rsim}} demonstrate the impact of vector space preprocessing (only L2-normalization versus L2 + mean centering + L2) on the RSIM isomorphism scores in different training regimes for \textsc{en-ar} and \textsc{en-ja} experiments.

\item {\bf Table~\ref{tab:evs-gh}} provides additional isomorphism scores, not reported in the paper, again stressing the importance of monolingual vector space preprocessing before learning any cross-lingual mapping.
\end{itemize}

\noindent (The actual tables and figures start on the next page for clarity.)

\begin{table*}[t]
\def\arraystretch{0.97}
\centering
{\footnotesize
\begin{tabularx}{\linewidth}{l Y Y Y Y Y}
\toprule
\multicolumn{6}{c}{\bf Full EN Vector Space -- Full ES Vector Space} \\
\midrule
{} & \multicolumn{1}{c}{HFREQ} & \multicolumn{1}{c}{MFREQ} & \multicolumn{1}{c}{LFREQ} & PANLEX & MUSE \\
\cmidrule(lr){2-2} \cmidrule(lr){3-3} \cmidrule(lr){4-4} \cmidrule(lr){5-5} \cmidrule(lr){6-6}
{BLI: Supervised (1k)} & {0.733} & {0.631}  & {0.621} & {0.448}  & {0.489}  \\
{BLI: Supervised+SL (1k)} & {0.774} & {0.711}  & {0.695} & {0.492}  & {0.536}  \\
{BLI: Supervised (5k)} & {0.759} & {0.685}  & {0.658} & {0.490}  & {0.533}  \\
{BLI: Supervised+SL (5k)} & {0.778} & {0.704}  & {0.691} & {0.491}  & {0.538}  \\
\hdashline
{RSIM} & {0.652} & {0.633}  & {0.559} & {--}  & {--}  \\
{Gromov-Hausdorff} & {0.274} & {0.208}  & {0.205} & {--}  & {--}  \\
{Eigenvector Similarity} & {4.86} & {6.32}  & {10.95} & {--}  & {--} \\
\bottomrule
\end{tabularx}
}
\vspace{-1.5mm}
\caption{Reference BLI (MRR reported) and isomorphism scores (all three measures discussed in the main paper are reported, computed on L2-normalised vectors) in a setting where we \textit{fully} train both English and Spanish monolingual vector spaces (i.e., training lasts for 15 epochs for both languages) on the \textit{full} data, without taking snapshots at earlier stages, and without data reduction simulation experiments.}
\label{tab:full-training}
\end{table*}

\begin{table*}[t]
\def\arraystretch{0.95}
\centering
{\footnotesize
\begin{tabularx}{\linewidth}{l YY YY YY YY}
\toprule
& \multicolumn{4}{c}{EN (full) - ES (snapshot)} & \multicolumn{4}{c}{EN (snapshot) - ES (full)} \\
\cmidrule(lr){2-5} \cmidrule(lr){6-9}
 & \multicolumn{2}{c}{EVS} & \multicolumn{2}{c}{GH} & \multicolumn{2}{c}{EVS} & \multicolumn{2}{c}{GH} \\
\cmidrule(lr){2-3} \cmidrule(lr){4-5} \cmidrule(lr){6-7} \cmidrule(lr){8-9}
{\# Updates} & {UNNORM} & {L2+MC+L2} & {UNNORM} & {L2+MC+L2} & {UNNORM} & {L2+MC+L2} & {UNNORM} & {L2+MC+L2} \\
\cmidrule(lr){2-2} \cmidrule(lr){3-3} \cmidrule(lr){4-4} \cmidrule(lr){5-5} \cmidrule(lr){6-6} \cmidrule(lr){7-7} \cmidrule(lr){8-8} \cmidrule(lr){9-9}
{100M} & {89.1 [42.7]} & {6.46 [5.88]} & {3.03 [3.19]} & {0.22 [0.30]} & {43.4 [20.9]} & {6.74 [26.3]} & {1.91 [3.66]} & {0.23 [0.33]} \\
{200M} & {162 [24.5]} & {3.67 [2.86]} & {3.45 [2.16]} & {0.25 [0.31]} & {38.3 [50.0]} & {9.15 [18.0]} & {2.51 [3.39]} & {0.27 [0.38]} \\
{600M} & {235 [26.2]} & {7.47 [3.13]} & {3.54 [1.80]} & {0.27 [0.29]} & {26.1 [20.0]} & {4.51 [15.0]} & {3.50 [2.21]} & {0.34 [0.37]} \\
{1B} & {125 [21.3]} & {4.69 [5.90]} & {3.31 [2.32]} & {0.24 [0.28]} & {45.2 [32.4]} & {12.5 [13.7]} & {3.35 [1.49]} & {0.27 [0.30]} \\
{2B} & {252 [23.4]} & {5.88 [6.44]} & {3.03 [2.13]} & {0.21 [0.31]} & {25.4 [9.43]} & {16.5 [10.1]} & {3.71 [2.03]} & {0.25 [0.29]} \\
{4B} & {360 [21.0]} & {8.78 [7.28]} & {2.54 [2.55]} & {0.15 [0.29]} & {141 [22.5]} & {5.33 [7.99]} & {3.04 [1.37]} & {0.18 [0.31]} \\
{6B} & {411 [16.1]} & {6.96 [8.22]} & {1.32 [2.22]} & {0.15 [0.23]} & {191 [16.7]} & {8.98 [11.4]} & {3.22 [0.99]} & {0.15 [0.37]} \\
\bottomrule
\end{tabularx}
}
\vspace{-1.5mm}
\caption{Eigen Vector Similarity (EVS) and Gromov-Hausdorff distance (GH) distance scores with two different monolingual vector space preprocessing strategies: (a) no normalisation at all (UNNORM); (c) L2-normalisation followed by mean centering (MC) and another L2-normalisation step, done as standard preprocessing in the VecMap framework \cite{Artetxe2018} (L2+MC+L2). We show the scores in relation to training duration (provided in the number of updates, i.e., \text{seen word tokens}), taking snapshots of the English or the Spanish vector space, and aligning it to a fully trained space on the other side. We show scores on HFREQ and [LFREQ] sets; lower is better (i.e., ``more isomorphic'').}
\label{tab:evs-gh}
\end{table*}

\section{Reproducibility: Data and Code}
\begin{itemize}
    \item Polyglot Wikipedias are available at: \\ {\small \url{https://sites.google.com/site/rmyeid/projects/polyglot}}
    \item Used fastText code is available at: \\ {\small
    \url{https://github.com/facebookresearch/fastText}}
    \item MUSE dictionaries: \\ {\small
    \url{https://github.com/facebookresearch/MUSE/tree/master/data}}
    \item VecMap framework accessible at: \\ {\small
    \url{https://github.com/artetxem/vecmap}}
    \item Iterative normalisation code available at: \\ {\small
    \url{https://github.com/zhangmozhi/iternorm}}
    \item Bilingual lexicons used in the study: \\ {\small
    \url{https://github.com/cambridgeltl/iso-study}}
    \item {Subsampled corpora for all languages:} \\ {\small
    \url{https://github.com/cambridgeltl/iso-study}}
\end{itemize}

For BLI, we use evaluation scripts from \newcite{Glavas2019}, available here: \url{https://github.com/codogogo/xling-eval}.

Our (research) code available for computing near-isomorphism via RSIM, EVS, and GH is also hosted online at: \url{https://github.com/cambridgeltl/iso-study}. The code relies on standard Python's stack for scientific computing (e.g., it uses \texttt{numpy}, \texttt{scipy}, \texttt{scikit-learn}, \texttt{networkx}). 



\end{document}